\documentclass[acmtog,nonacm,screen]{acmart}

\usepackage{subcaption}

\AtBeginDocument{%
  \providecommand\BibTeX{{%
    \normalfont B\kern-0.5em{\scshape i\kern-0.25em b}\kern-0.8em\TeX}}}

\setcopyright{acmcopyright}
\copyrightyear{2018}
\acmYear{2018}
\acmDOI{XXXXXXX.XXXXXXX}

\acmConference[Conference acronym 'XX]{Make sure to enter the correct
  conference title from your rights confirmation emai}{June 03--05,
  2018}{Woodstock, NY}
\acmPrice{15.00}
\acmISBN{978-1-4503-XXXX-X/18/06}

\newcommand{\ad}[1]{\textcolor{blue}{[AD: #1]}}

\usepackage{nicefrac}
\usepackage{xfrac}
\usepackage{enumitem}
\usepackage{bm}
\usepackage{lettrine}

\begin{document}

\title{NeST: Neural Stress Tensor Tomography by leveraging 3D Photoelasticity}

\author{Akshat Dave}
\affiliation{
\institution{Massachusetts Institute of Technology Media Lab}
\city{Cambridge}
\state{MA}
\country{USA}
}
\email{ad74@media.mit.edu}
\author{Tianyi Zhang}
\authornote{Equal Contribution}
\affiliation{
\institution{Rice University}
\city{Houston}
\state{TX}
\country{USA}
}
\author{Aaron Young}
\authornotemark[1]
\affiliation{
\institution{Massachusetts Institute of Technology Media Lab}
\city{Cambridge}
\state{MA}
\country{USA}
}
\author{Ramesh Raskar}
\affiliation{
\institution{Massachusetts Institute of Technology Media Lab}
\city{Cambridge}
\state{MA}
\country{USA}
}
\author{Wolfgang Heidrich}
\affiliation{
\institution{King Abdullah University of Science and Technology}
\country{Saudi Arabia}
}
\author{Ashok Veeraraghavan}
\affiliation{
\institution{Rice University}
\city{Houston}
\state{TX}
\country{USA}
}

\renewcommand{\shortauthors}{Dave et al.}

\begin{abstract}
Photoelasticity enables full-field stress analysis in transparent objects through stress-induced birefringence. Existing techniques are limited to 2D slices and require destructively slicing the object. Recovering the internal 3D stress distribution of the entire object is challenging as it involves solving a tensor tomography problem and handling phase wrapping ambiguities. We introduce NeST, an analysis-by-synthesis approach for reconstructing 3D stress tensor fields as neural implicit representations from polarization measurements. Our key insight is to jointly handle phase unwrapping and tensor tomography using a differentiable forward model based on Jones calculus. Our non-linear model faithfully matches real captures, unlike prior linear approximations. We develop an experimental multi-axis polariscope setup to capture 3D photoelasticity and experimentally demonstrate that NeST reconstructs the internal stress distribution for objects with varying shape and force conditions. Additionally, we showcase novel applications in stress analysis, such as visualizing photoelastic fringes by virtually slicing the object and viewing photoelastic fringes from unseen viewpoints. NeST paves the way for scalable non-destructive 3D photoelastic analysis.
\end{abstract}


\begin{teaserfigure}
  \includegraphics[width=\textwidth]{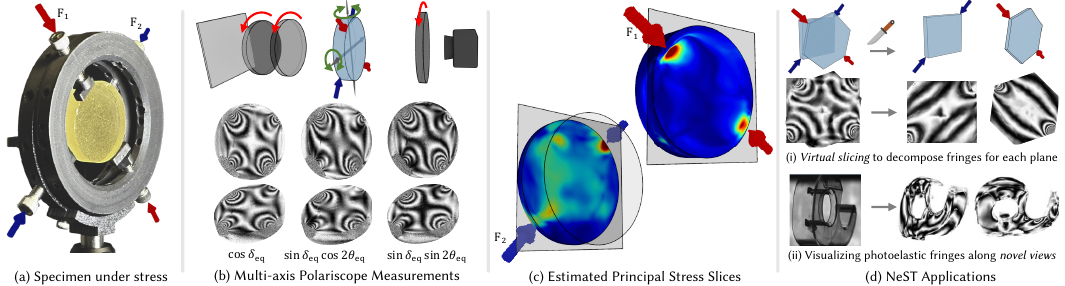}
  \caption{\textbf{NeST Overview.} Transparent objects, when subject to external forces (a), exhibit birefringence that manifests as fringes under polarized illumination. We present NeST, an approach to leverage this 3D photoelasticity through a multi-axis polariscope setup (b). We develop an analysis-by-synthesis approach using neural implicit fields to non-desctructively reconstruct the internal 3D stress distribution in the object from the captured measurements (c). NeST also enables new ways of visualizing the underlying 3D stress by rendering the photoelastic fringes obtained by interactively slicing or rotating the object. }
  \label{fig:teaser}
\end{teaserfigure}


\maketitle
\section{Introduction}

Photoelasticity is an optical phenomenon that allows full-field observation and quantification of stress distributions in transparent materials. When an object is subject to mechanical load, it exhibits birefringence causing the orthogonal polarization states of transmitted light to have a relative phase retardation. This phase retardation visually manifests as fringes when the object is placed between polarizers, with denser fringes indicating higher stress concentration. Photoelasticity has broad applications ranging from stress analysis of dental implants \cite{goiato2014photoelastic,ramesh2016digital} to quality control of glass panels \cite{kasper2016quality}. 


Most existing photoelasticity techniques operate on 2D slices or projections of a 3D object \cite{ramesh2021developments}. In a planar slice, the distribution of photoelastic fringes is related directly to the distribution of principal stress difference. The most common approach for 3D analysis involves stress freezing: locking in the stress distribution in the object by temperature cycling. 
Thin slices are then manually cut. 2D stress fields are then obtained for each of the slices, which when put together create a 3D stress field. Unfortunately, the entire process is expensive, time-consuming and most importantly destructive.


3D photoelasticity \cite{aben1979integrated} aims to recover the underlying 3-dimensional stress distribution of the \textit{entire} object without the need for physical 2D slicing thereby scaling photoelasticity to more unstructured and non-destructive scenarios. The polarization state of each light ray encodes the stress distribution along its path through the object and is modeled as an equivalent Jones matrix. This Jones matrix can be measured using controlled polarization illumination and detection \cite{collett2005field}. 

Reconstructing the complete 3D stress tensor field from integrated polarization measurements is challenging for two reasons. First, the stress at each point is a $3\times3$ tensor, but each ray only encodes a $2\times 2$ projection. Multiple object or camera rotations are required to reconstruct the tensor field, causing this \textit{tensor tomography} problem to require more measurement diversity than conventional scalar tomography like X-ray CT \cite{szotten2011limited}. Second, the relative phase difference between the polarization states that encodes the stress information is always wrapped between $0$ and $2\pi$. Thus the captured polarization measurements require \textit{phase unwrapping} to recover the full stress information.

Existing approaches aim to tackle these challenges of phase unwrapping and tensor tomography in two \textit{separate} steps \cite{abrego2019experimental}. First, the polarization measurements for each rotation are individually phase unwrapped using techniques extended from single-view 2D photoelasticity \cite{tomlinson2002use}. Then the unwrapped measurements are approximated as a linear projection of the underlying stress tensor field \cite{sharafutdinov2012integral}. We demonstrate through experimental captures that this two-step fails under realistic stress distributions. Separately phase unwrapping measurements for each rotation do not provide a way to enforce consistency across iterations resulting in artifacts in the unwrapped phase (Fig.~\ref{fig:pipeline}). Furthermore, we observe that the linear tensor tomography model does not match the captures under realistic scenarios (Fig.~\ref{fig:pipeline}).

Our key insight is to \textit{jointly} handle phase unwrapping and tensor tomography. We develop a differentiable Jones calculus-based forward model that maps the underlying 3D stress tensor distribution to the captured polarization measurements. The existing linear tensor tomography model \cite{sharafutdinov2012integral} is a first-order approximation of our general non-linear forward model and our model can faithfully explain real measurements (Fig.~\ref{fig:pipeline}). With this model, we present NeST, an analysis-by-synthesis approach to reconstruct full-field 3D stress tensor distribution directly from captured intensity measurements. 
 Inspired by the recent advancements in inverse neural rendering, we employ neural implicit representations for the unknown stress tensor field. Neural representations provide computationally efficiency and adaptive sampling in representing and reconstructing highly concentrated stress fields.  
 
We develop a multi-axis polariscope hardware setup to experimentally validate our approach (Fig.~\ref{fig:acquisition_setup}). This setup involves multiple measurements of the object under yaw-pitch rotation (multi-axis) and rotation of the polarizing elements (polariscope). NeST can reconstuct internal stress distribution for objects with a variety of 3D shapes and loading conditions (Fig.~\ref{fig:real_qualitative_disk}). By neural rendering the estimated internal stress, NeST enables novel approaches to visualize the stress tensor distribution (Fig.~\ref{fig:app_virtual_slicing},\ref{fig:app_visualization}).  We also qualitatively validate (Fig.~\ref{fig:sim_showcase}) and analyze our approach (Fig.~\ref{fig:sim_analysis_coeff},\ref{fig:sim_analysis_rotation}) through a simulated dataset of common stress distributions. 
\paragraph{Our Contributions}
To summarize, we demonstrate the following:
\begin{itemize}[leftmargin=*]
    \item \textbf{Differentiable non-linear forward model} for 3D photoelasticity that faithfully matches the captured polarization measurements. 
    \item \textbf{Neural analysis-by-synthesis approach} that reconstructs internal stress distribution as a neural implicit representation from captured polarization measurements.
    \item \textbf{Experimental validation} of proposed stress tensor tomography through a multi-axis polariscope setup.
    \item \textbf{Simulated and real-world datasets} of 3D photoelastic measurements on a variety of object and load geometries. 
    \item \textbf{Novel stress visualizations} from the learned neural stress fields such as visualizing photo-elastic fringes obtained by virtually slicing the object and viewing the object along unseen views.  
\end{itemize}
The codebase and datasets will be made public upon acceptance.

\paragraph{Scope}
Although our work presents a major advance in 3D stress analysis, several limitations remain. First, we do not currently model absorption, or reflection and refraction at object boundaries. The latter could be handled by refractive index matching, or by pre-scanning the geometry of the object, and ray-tracing the refracted ray path at object boundaries (see Sec~\ref{sec:discussion} for a more detailed discussion). Since our method requires many images, static geometry and loading conditions are required. Finally, our method shares all the inherent limitations of photoelasticity methods, i.e. the object needs to be made of a transparent medium, and the deformation needs to be in the elastic regime.

\section{Related Work}
\begin{figure*}[ht]
    \centering
    \includegraphics[width=\textwidth]{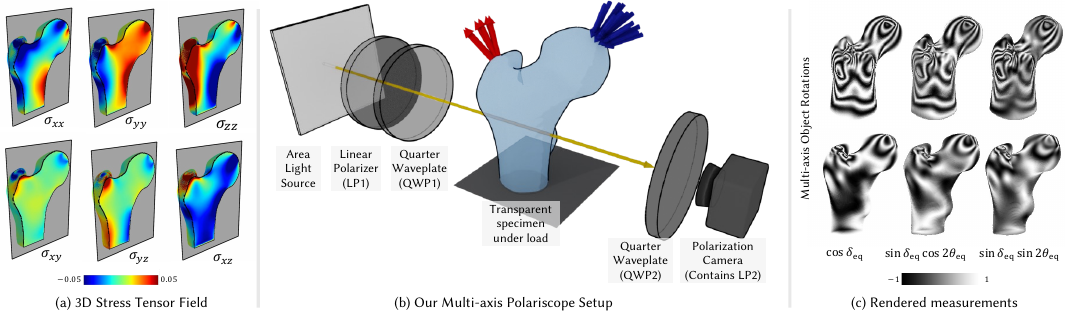}
    \caption{\textbf{Overview of our formulation.} We develop a formulation to render stress-induced birefringence from the stress tensor field distribution (a) in a 3D transparent object. The stress field at each point changes the polarization state of a ray passing through that point (Sec.~\ref{sec:birefringence}, Fig.~\ref{fig:projected_stress_tensor}). We integrate these polarization changes for all points along the ray to obtain an equivalent Jones matrix (Sec.~\ref{sec:integrated_photoelasticity}, Fig.~\ref{fig:equivalence_theorem}). These Jones matrices and consequently the underlying stress tensors are then measured through our multi-axis polariscope capture setup (b) (Sec.~\ref{sec:polariscope}, Fig.~\ref{fig:render_multiple_views}). In (c), we visualize the multi-axis polariscope measurements rendered from our formulation for the stress field in (a). 
    }
    \label{fig:cir_pol}
\end{figure*}

\subsection{Polarimetric Imaging}
\paragraph{Applications in vision and graphics} Polarization characterizes the direction of oscillation of light waves \cite{collett2005field} and encodes useful scene properties. There has been significant progress in exploiting polarization cues for graphics and vision applications, including reflectance separation \cite{lyu2019reflection, li2020reflection},  material segmentation \cite{kalra2020deep, mei2022glass}, navigation \cite{yang2018polarimetric}, dehazing \cite{schechner2001instant}, shape estimation \cite{chen2022perspective, kadambi2015polarized, cui2017polarimetric, lei2022shape, tozza2017linear, zhao2022polarimetric} and appearance capture \cite{deschaintre2021deep, riviere2017polarization, ghosh2011multiview, ngo2015shape, baek2018simultaneous, hwang2022sparse, ghosh2010circularly, dave2022pandora}. 

\paragraph{Birefringence} In this work, we leverage the polarization
phenomenon of birefringence that is relatively underexplored by the
vision and graphics community. Birefringence is an optical property in
which the refractive index depends on the polarization and propagation
direction of light. It occurs in optically anisotropic materials where
the structure or stresses induce different indices along different
axes. Birefringence has been widely studied and utilized for
mechanical stress analysis via photoelasticity techniques
\cite{ramesh2021developments}. Birefringence is also exploited for imaging fibrous tissues
\cite{huang2002linear}, cancer pathology~\cite{ushenko2013complex},
and liquid crystal displays~\cite{yeh2009optics}. Multi-layer liquid
crystal displays~\cite{lanman2011polarization} use a tomographic
polarization model for generating 3D images that however neglects
birefringence.  Our work focuses on leveraging birefringence for full
3D stress measurement via novel neural tomography approaches.

\subsection{Photoelasticity}
\paragraph{2D Photoelasticity}
Photoelasticity is an optical phenomenon based on stress-induced birefringence in transparent objects. It has been extensively utilized for full-field stress analysis in various fields such as structural engineering \cite{scafidi2015review,ju_quantitative_2018}, material science \cite{wang_mechanical_2017,ju_photoelastic_2018,ju_quantification_2019} and biomechanics \cite{joseph2015imaging,tomlinson_photoelastic_2015,falconer_developing_2019,doyle_use_2012,sugita_photoelasticity-based_2019}. Coker, Filon and Frocht detailed the core principles and methodologies of photoelasticity in their seminal books \cite{frocht_photoelasticity_1941,coker1957treatise}, while Dally et al.~\shortcite{dally1978experimental} applied these techniques to engineering problems. The emergence of digital photography \cite{ramesh_digital_2011,kulkarni_optical_2016} and RGB cameras \cite{ajovalasit_review_2015} has significantly advanced the field. Phase shifting technique \cite{patterson1991towards} emerged as a practical approach to recover stress distribution from photoelastic fringes by rotating polariscope elements. Recently, there has been progress in applying machine learning techniques for estimating stress distribution from photoelastic fringes \cite{brinez2024deep,brinez2022photoelastnet,lin2024pinn}. These works aim to recover 2D stress distributions in planar objects, while we focus on the more challenging scenario of 3D stress distributions from 3D photoelastic measurements.

\paragraph{3D Photoelasticity}
Extending photoelasticity to 3D objects traditionally involved freezing the stress distribution using temperature cycling and then manually slicing the object into sections for analysis \cite{cernosek_three-dimensional_1980}, but this process is costly, time-consuming, and destructive. 3D photoelasticity was conceptualized to overcome these limitations and transition photoelastic stress analysis from 2D slices to reconstructing full 3D stress tensor fields within objects \cite{theocaris_three-dimensional_1979,orourke_threedimensional_1951,weller_three-dimensional_1941}. Aben~\shortcite{aben1979integrated} formalized the concept of integrated photoelasticity that models continuous integration of stress variation along a ray through the object. Bussler et al.~\shortcite{bussler2015photoelasticity} developed a framework to render photoelastic fringes from 3D stress distributions by solving for integrated photoelasticity using Runge Kutta numerical integration  They focus on rendering photoelastic fringes from known stress distribution for visualization purposes. Their numerical integration-based forward model is not differentiable and cannot be used for analysis-by-synthesis techniques. We develop a differentiable forward model for 3D photoelasticity that enables 3D stress reconstruction by neural analysis-by-synthesis techniques. In supplement, we demonstrate how our framework faithfully approximates the integrated photoelasticity model under sufficiently low render step size.

\paragraph{Photoelastic Tomography}
There has been very limited work on directly acquiring and reconstructing 3D photoelasticity. Our paper presents a major advance in this direction. Most of existing works approximate the photoelastic forward model to be linear and pose the stress field reconstruction as a linear tensor tomography problem \cite{sharafutdinov2012integral}. Sharafutdinov et al.~\shortcite{sharafutdinov2012integral} and Aben et al.~\shortcite{aben2005photoelastic} formulate the reconstruction of a single tensor element, while Hammer et al.~\shortcite{hammer2005reconstruction} propose a linear tomography approach to reconstruct all the tensor elements that is developed to handle incomplete data in Lionheart et al.~\shortcite{lionheart2009reconstruction}. Szotten \shortcite{szotten2011limited} showed simulated results with the Lionheart et al.~\shortcite{lionheart2009reconstruction} technique and condition the problem using Hilbert transform methods. Abrego~\shortcite{abrego2019experimental} developed an experimental framework to test the algorithms developed by Szotten~\shortcite{szotten2011limited} and concluded that their technique is unable to scale to real experimental data. We demonstrate real experimental results that agree with these findings. We demonstrate in Fig.~\ref{fig:pipeline} that linear model is incapable of explaining experimental photoelastic measurements especially when the stress variation is large. We develop a differentiable non-linear forward model and analysis-by-synthesis technique that faithfully explains the experimental measurements and reconstructs the underlying stress variation. 

\begin{figure*}[!h]
    \centering
    \includegraphics[width=\linewidth]{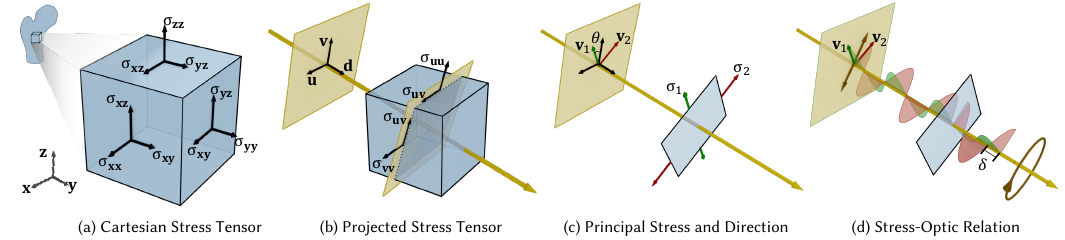}
    \caption{\textbf{Modeling stress induced birefringence.} Each point in the specimen under stress corresponds to a $3 \times 3$ Cartesian stress tensor (a). A ray passing through this point encodes information about the $2 \times 2$ projection of the stress tensor on a plane perpendicular to the ray (b). The projected stress tensor can be represented by principle stress directions corresponding to solely normal stress (c). Difference in principle stress values results in a phase difference that corresponds to a change in the polarization of the light ray (d).  }
    \label{fig:projected_stress_tensor}
\end{figure*}



\subsection{Neural Fields and Neural Rendering}
Implicit neural representations (INRs) have emerged as a powerful technique to represent 3D objects using trainable models, such as multi-layer perceptrons (MLPs) \cite{mildenhall2021nerf}. These models learn the mapping from spatial coordinates to object properties, like density or color. INRs offer several advantages: continuous representation in spatial coordinates, decoupling of object complexity from sampling resolution, memory efficiency, and a compact representation that acts as a sparsity prior, aiding inverse problems. Furthermore, INRs can be combined with differentiable forward simulators to solve inverse problems via gradient backpropagation.
While initially developed for multiview stereo reconstruction \cite{mildenhall2021nerf}, INRs have found diverse applications in representing and reconstructing field data \cite{xie2022neural}. These include fitting 2D images \cite{sitzmann2020implicit}, representing 3D geometric properties like occupancy fields and signed distance functions (SDFs) \cite{sitzmann2020implicit,yariv2021volume,wang2021neus}, and modeling various signal or field quantities in scientific and medical fields, such as geodesy \cite{izzo2021geodesy}, black hole imaging \cite{levis2024orbital}, audio signals \cite{gao2021objectfolder}, protein structures \cite{zhong2019reconstructing}, computed tomography \cite{corona2022mednerf}, MRI \cite{shen2022nerp}, ultrasound imaging \cite{wysocki2024ultra}, and synthetic aperture sonar \cite{reed2023neural,reed2021implicit}. NeST paves the way for neural fields-based approaches in the field of non-destructive stress analysis. 

\section{Background}
\subsection{Jones Calculus}
Jones calculus is a mathematical formalism used to describe the polarization state of light. It represents the polarization state as a $2\times1$ Jones vector $\mathbf{E}$ containing complex components that denote the amplitude and phase of two orthogonal polarization modes.
\begin{align}    
\mathbf{E} = \begin{pmatrix} E_x\ E_y\end{pmatrix}
\label{eq:Ex_Ey}
\end{align}
where $E_x$ and $E_y$ are the x and y components of the electric field vector.
Under Jones calculus, the effect of an optical element on the polarization can be represented as a $2\times2$ Jones matrix operating on the input Jones vector, 
\begin{align}    
\mathbf{E}_\text{out} = \mathbf{J} \mathbf{E}_\text{in} \; ,
\end{align}

where $\mathbf{J}$ is the Jones matrix of the optical element, $\mathbf{E}_\text{in}$ is the input Jones vector, and $\mathbf{E}_\text{out}$ is the output Jones vector.

\subsection{Stress Tensor Field}
Consider an object of arbitrary shape subject to external mechanical forces. These external forces would result in a three-dimensional distribution of body forces throughout the body of the object modeled as mechanical stress. 

The stress at a point $\mathbf{p}$ in the material is commonly represented by the second-order Cartesian stress tensor \cite{dally1978experimental}. Considering a small, axis-aligned cubic element at $\mathbf{p}$, the stress tensor characterizes the forces on each face of this cube projected along the $x$, $y$ and $z$ axis.  The stress tensor, $S$, can be expressed as a matrix with rows as the $x$, $y$, $z$ normal directions of the faces and columns as the forces along $x$, $y$, $z$ directions for each face.
\begin{equation}
\mathbf{S}(\mathbf{p}) = 
\begin{bmatrix}
\sigma_{xx} & \sigma_{xy} & \sigma_{xz} \\
\sigma_{yx} & \sigma_{yy} & \sigma_{yz} \\
\sigma_{zx} & \sigma_{zy} & \sigma_{zz}
\end{bmatrix}
\end{equation} 

Under the conditions of equilibrium, $\sigma_{ij} = \sigma_{ji}$. Thus at each point, the stress tensor can be expressed as a symmetric matrix with six unkowns:
\begin{equation}
\mathbf{S}(\mathbf{p}) = 
\begin{bmatrix}
\sigma_{xx} & \sigma_{xy} & \sigma_{xz} \\
\sigma_{xy} & \sigma_{yy} & \sigma_{yz} \\
\sigma_{xz} & \sigma_{yz} & \sigma_{zz}
\end{bmatrix}
\label{eq:stress_tensor}
\end{equation}
The principal stresses at a point are defined as the normal stresses calculated on planes where the shear stresses are zero. These principal stresses can be obtained by performing an eigendecomposition of the Cartesian stress tensor in Eq.~\ref{eq:stress_tensor}. The magnitudes of the major, middle, and minor principal stresses are given by the eigenvalues of the stress tensor, sorted from highest to lowest. The corresponding eigenvector represents the directions of the principal stress.
\section{Photoelastic Image Formation Model}
\label{sec:image_formation_model}

In this section, we derive how the polarimetric light transport through a transparent object encodes its underlying three-dimensional stress tensor distribution. We demonstrate how stress present at a certain point within the object induces birefringence that we model through Jones calculus. Then we present a volume rendering approach to integrate the stress-birefringence effects into an equivalent Jones matrix that we can measure with a multi-axis polariscope setup. 

\subsection{Stress-induced Birefringence}
\label{sec:birefringence}
Consider a ray with origin $\mathbf{o}$ and direction $\mathbf{d}$ propagating through the transparent object under stress. The points $\mathbf{p}(t)$ along the ray and within the object are parameterized as:
\begin{equation}
    \mathbf{p}(t) = \mathbf{o} + t\mathbf{d} \quad t \in [ t_n, t_f] \;.
\end{equation} 
The stress at the point $\mathbf{p}(t)$ is modeled by the Cartesian stress tensor $\mathbf{S}(t)$ (Eq.~\ref{eq:stress_tensor}). We demonstrate how, due to stress-induced birefringence, this stress tensor is encoded in the polarimetric light transport using Jones calculus. 

\paragraph{Projected stress tensor}
Consider the plane orthogonal to the ray $(\mathbf{o}, \mathbf{d})$ is spanned by two orthonormal basis vectors, $\mathbf{u}$ and $\mathbf{v}$. The projection of the $3\times 3$ stress tensor $\mathbf{S}(t)$ along the orthogonal plane with $\mathbf{u} - \mathbf{v}$ axes is denoted as the $2 \times 2$ symmetric matrix $\mathbf{S}'(t)$ 
\begin{equation}
   \mathbf{S}'(t) = \begin{bmatrix} \mathbf{u}^T \mathbf{S}(t)\mathbf{u} & \mathbf{u}^T\mathbf{S}(t)\mathbf{v} \\ \mathbf{u}^T\mathbf{S}(t)\mathbf{v} & \mathbf{v}^T\mathbf{S}(t)\mathbf{v} \end{bmatrix} 
   \triangleq \begin{bmatrix}
      \sigma_{\mathbf{uu}} & \sigma_{\mathbf{u}\mathbf{v}} \\
      \sigma_{\mathbf{uv}} & \sigma_{\mathbf{vv}}
   \end{bmatrix} \; . 
   \label{eq:projected_stress}
\end{equation}
\paragraph{Principal stresses}
For general choices of the orthonormal basis $(\mathbf{u},\mathbf{v})$, $\mathbf{S}'(t)$ is a non-diagonal matrix, with the diagonal entries corresponding to the normal stress and non-diagonal entries corresponding to tangential stress. The major and minor principal stress directions, $(\mathbf{w_1},\mathbf{w_2})$, are defined as an orthonormal basis that results in a diagonal projected stress $\mathbf{S'}$, i.e.,
\begin{equation}
 \begin{bmatrix} \mathbf{w}_1^T \mathbf{S}(t)\mathbf{w}_1 & \mathbf{w}_1^T\mathbf{S}(t)\mathbf{w}_2 \\ \mathbf{w}_1^T\mathbf{S}(t)\mathbf{w}_2 & \mathbf{w}_2^T\mathbf{S}(t)\mathbf{w}_2 \end{bmatrix} 
   \triangleq \begin{bmatrix}
      \sigma_1 & 0 \\
      0 & \sigma_2
   \end{bmatrix} \; ,
   \label{eq:principal_stress}
\end{equation}
with $\sigma_1 > \sigma_2$.
Along $\mathbf{w}_1$ and $\mathbf{w}_2$ directions, there is only normal stress: $\sigma_1$ and $\sigma_2$ respectively which are termed as the major and minor principal stresses. In the supplement, we derive that for any general $\mathbf{u},\mathbf{v}$ basis the difference of principle stresses depends on the components of $\mathbf{S}'$  as 
\begin{equation}
   \sigma_2 - \sigma_1 = \sqrt{\left(\sigma_{\mathbf{vv}} - \sigma_{\mathbf{uu}} \right)^2 + {\sigma_{\mathbf{uv}}}^2} \; .
\end{equation}
The angle $\theta$ made by the principle stress $\sigma_1$ with $\mathbf{u}$ is 
\begin{equation}
   \theta = \frac{1}{2}\tan^{-1}\left(\frac{2\sigma_{\mathbf{uv}}}{\sigma_{\mathbf{vv}}-\sigma_{\mathbf{uu}}}\right) \; . \label{eq:theta}
\end{equation}
\paragraph{Stress-optic relation}
Consider the segment of length $\Delta$ along the ray around the point $\mathbf{p}(t)$. The length $\Delta$ is small so that there is no stress variation along the segment. Assuming weak birefringence, stress-optic relation \cite{dally1978experimental} states that the phase difference $\delta(t)$ between the principal directions $(\mathbf{v_1}$,$\mathbf{v_2})$ is proportional to principal stress difference $\sigma_1 - \sigma_2$  and is given as:
\begin{equation}
    \delta(t) = \frac{2\pi \Delta C}{\lambda} \left( \sigma_2 - \sigma_1 \right),
    \label{eq:stress_optic}
\end{equation}
where $C$ is the stress-optic coefficient that depends on the material properties of the object and $\lambda$ is the wavelength of light.
\paragraph{Jones matrix}
Next we can model this stress-induced birefringence at the small segment around $\mathbf{p}(t)$ as a Jones matrix denoted as $\mathbf{J}(t)$. The definition of Jones matrix depends on the orthonal basis for Jones vectors. If we consider the Jones vectors are defined based on principal stress directions $(\mathbf{v_1},\mathbf{v_2})$, then the Jones matrix $\hat{\mathbf{J}}(t)$ corresponds to that of a retarder with retardance $\delta$ and slow axis along $\mathbf{v_1}$ which is given as \cite{collett2005field}, 
\begin{align}
    \hat{\mathbf{J}}(t) = \begin{bmatrix}
                            e^{i\nicefrac{\delta}{2}} & 0 \\
                            0 & e^{-i\nicefrac{\delta}{2}} \\ 
                          \end{bmatrix} \; .
\end{align}
The principal directions are not known a priori. The general orthonormal basis $(\mathbf{u},\mathbf{v})$ is rotated $\theta$ from the principal directions $(\mathbf{v_1},\mathbf{v_2})$ as given by Eq.~\ref{eq:theta}. For the Jones vectors with orthonormal basis along $(\mathbf{u},\mathbf{v})$, the Jones matrix corresponds to that of a retarder with phase $\delta$ and orientation $\theta$ from the slow axis ($\mathbf{u}$) which is given as \cite{collett2005field}:
\begin{align}
\mathbf{J}(t) = \cos \left(\nicefrac{\delta}{2}\right)
\begin{bmatrix}
1 & 0 \\
0 & 1
\end{bmatrix}
+ i \sin \left(\nicefrac{\delta}{2}\right)
\begin{bmatrix}
\cos 2\theta & \sin 2\theta \\
\sin 2\theta & -\cos 2\theta
\end{bmatrix}
\label{eq:J_main} \; ,
\end{align}

\subsection{3D Photoelasticity}
\label{sec:integrated_photoelasticity}

\paragraph{Integrated photoelasticity using ray marching}
Consider $N$ samples along the ray parameterized as $t_i, i = 1, 2, \ldots, N$. Considering sufficiently large number of samples, we approximate the stress between the samples \( t_i \) and \( t_{i+1} \) to be constant., From Eq. ~\ref{eq:J_main}, the segment between $t_i$ and $t_{i+1}$ can then be approximated as a retarder with retardance $\delta_i$ and slow axis orientation $\theta_i$.

We can express the Jones matrix in terms of these parameters as
\begin{equation}
\mathbf{J}\left(t_i\right) = \cos \left(\nicefrac{\delta_i}{2}\right)
\begin{bmatrix}
1 & 0 \\
0 & 1
\end{bmatrix}
+ i \sin \left(\nicefrac{\delta_i}{2}\right)
\begin{bmatrix}
\cos 2\theta_i & \sin 2\theta_i \\
\sin 2\theta_i & -\cos 2\theta_i
\end{bmatrix}
\label{eq:J_retarder} \; .
\end{equation} 
Aggregating along the entire ray, we can express the equivalent Jones vector $\mathbf{J}_\mathrm{eq}$ as the series of complex matrix multiplications
\begin{equation}
\mathbf{J}_\mathrm{eq} = \prod_{i=1}^{N} \mathbf{J}(t_i) \;.
\label{eq:integrated_photoelasticity_Jones}
\end{equation}

\begin{figure}
    \centering
    \includegraphics[width=\linewidth]{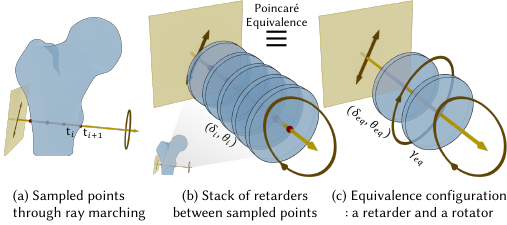}
    \caption{\textbf{Modeling integrated photoelasticity.} By ray marching, we sample points along the queried ray (a). We model each segment between the sampled points, $t_{i}$ and $t_{i+1}$ as a retarder with retardance $\delta_i$ and slow-axis orientation $\theta_i$ (b). From the Poincaré equivalence theorem, stack of any number of retarders is equivalent to a single retarder with parameters $(\delta_{\textit{eq}},\theta_{\textit{eq}})$ followed by a rotator with rotation $\gamma_{\mathrm{eq}}$ (c).}
    \label{fig:equivalence_theorem}
\end{figure}
\paragraph{Employing equivalence theorem} The net Jones matrix can be understood as a combination of $N$ retarders along the ray each with a different retardance $\delta_i$ and orientation $\theta_i$. Poincare's equivalence theorem states that the polarization properties of a combination of retarders are equivalent to that of a retarder followed by a rotator. Thus, the aggregate Jones matrix $\mathbf{J}_{\mathrm{eq}}$ can be represented by the product of Jones matrix of retarder with retardance $\delta_\mathrm{eq}$ and orientation $\theta_\mathrm{eq}$ and a rotator with $\gamma_\mathrm{eq}$.  The angles $\delta_\text{eq}$, $\theta_\text{eq}$ , and $\gamma_\text{eq}$  are denoted as the characteristic parameters \cite{aben1979integrated}. In the supplement, we derive the equivalence theorem for Eq.~\ref{eq:integrated_photoelasticity_Jones} and show that equivalent Jones matrix $\mathbf{J}_\text{eq}$ can be expressed based on the characteristic parameters as     
\begin{align}
\mathbf{J}_\text{eq} &= 
\cos{\left(\nicefrac{\delta_\text{eq}}{2}\right)}
\begin{bmatrix}
\cos 2\gamma_\text{eq} & \sin 2\gamma_\text{eq}\\
\sin 2\gamma_\text{eq} & -\cos 2\gamma_\text{eq} 
\end{bmatrix}  \nonumber \\
&+ i \sin{\left(\nicefrac{\delta_\text{eq}}{2}\right)} 
\begin{bmatrix}
\cos 2\theta'_\text{eq} & \sin 2\theta'_\text{eq}\\
\sin 2\theta'_\text{eq} & -\cos 2\theta'_\text{eq} 
\end{bmatrix} 
\label{eq:J_eq} \;,
\end{align}
where $\theta'_\text{eq} = \theta_\text{eq} - \gamma_\text{eq}$.
\begin{figure}
    \centering
    \includegraphics[width=\linewidth]{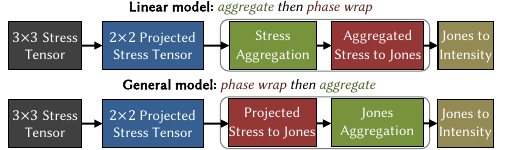}
    \caption{\textbf{General vs linear 3D photoelasticity model.} The proposed general 3D photoelasticity formulation involves converting projected stress tensors at each point on the ray to Jones matrices (\textit{first phase wrap}) that are then integrated using matrix multiplications (\textit{then aggregate}). We also derive a first-order approximation of this model that involves first summing the projected stress tensors (\textit{first aggregate}) and then converting the aggregated stress tensor to Jones matrix (\textit{then phase wrap})
    }
    \label{fig:general_vs_linear}
\end{figure}

\paragraph{Linear approximation} In the supplement, we show that under first-order approximations, our formulation simplifies to a linear combination of projected stress followed by phase wrapping. Prior works leverage this approximation to formulate photoelastic tomography as linear tensor tomography \cite{szotten2011limited,sharafutdinov2012integral}. Consider the projected stress components $\sigma_{\text{lin}}$ and $\tau_{\text{lin}}$ defined as
\begin{equation}
\sigma_{\text{lin}} = \sum_{i=1}^{N} (\sigma_\mathbf{uu}(t_i) - \sigma_\mathbf{vv}(t_i)) \Delta t_i
\quad \quad
\tau_{\text{lin}} = \sum_{i=1}^{N} \tau_\mathbf{vv}(t_i) \Delta t_i
\label{eq:sig_tau_lin}
\end{equation}
In the supplement, we also show that the aggregated Jones matrix under first-order approximation $\mathbf{J}_{\text{lin}}$ can be written as 
\begin{equation}
\mathbf{J}_{\text{lin}} = \cos \left(\nicefrac{\delta_{\text{lin}}}{2}\right) \mathbf{I} + j \sin \left(\nicefrac{\delta_{\text{lin}}}{2}\right)
\begin{bmatrix}
\cos 2\theta_{\text{lin}} & \sin 2\theta_{\text{lin}} \\
\sin 2\theta_{\text{lin}} & -\cos 2\theta_{\text{lin}}
\end{bmatrix}
\label{eq:J_approx} \;,
\end{equation}
where
\begin{align}
\delta_{\text{lin}} = -\frac{4\pi C}{\lambda} \sqrt{\sigma_{\text{lin}}^2 + \tau_{\text{lin}}^2} \quad \quad 
\theta_{\text{lin}} = \frac{1}{2} \tan^{-1} \left( \frac{\sigma_{\text{lin}}}{\tau_{\text{lin}}} \right)
\end{align}
\label{eq:delta_theta}
\paragraph{Our general non-linear model vs linear approximation} In the most general case, the projected stress tensor at each point is encoded as Jones matrices (Eq.~\ref{eq:J_eq}) where the stress values are phase wrapped due to the periodic nature of polarization. Then all Jones matrices are aggregated using Jones matrix multiplication. The linear approximation employed by prior works \cite{szotten2011limited,sharafutdinov2012integral} first aggregates the projected stress values using Eq.~\ref{eq:sig_tau_lin} and then performs phase wrapping (Eq.~\ref{eq:J_approx}). We summarize the differences between these models in Fig.~\ref{fig:general_vs_linear}.
Moreover, the aggregated Jones matrix in the general case (Eq.~\ref{eq:J_eq}) is parameterized by three characteristic parameters $\delta_{net}$, $\theta_{net}$ and $\gamma_{net}$. In the linear approximation, the aggregated Jones matrix (Eq.~\ref{eq:J_approx}) is modeled as an equivalent retarder with parameters $\delta_{lin}$ and $\theta_{net}$ and no rotator. Thus the linear approximation is not sufficient to model the rotation of polarization due to integrated photoelasticity. In Fig.~\ref{fig:pipeline}, we demonstrate how the linear model fails to explain the photoelastic fringes for a complicated stress field distribution.

\subsection{Multi-axis Polariscopy}
\label{sec:polariscope}

\begin{figure}[h]
    \centering
    \includegraphics[width=\linewidth]{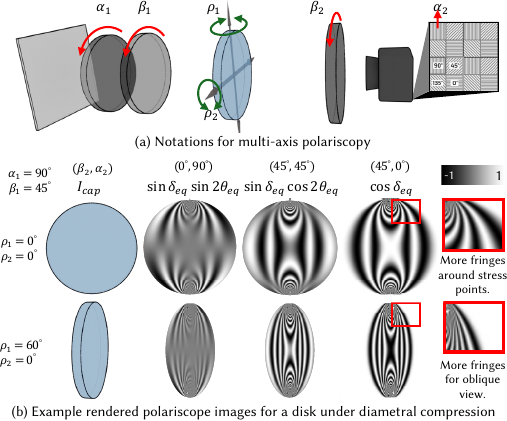}
    \caption{\textbf{Multi-axis polariscope capture setup} By varying yaw-pitch rotations of the specimen and varying rotations of quarter waveplates and polarizers (a), we capture the intensity measurements (b). These measurements exhibit fringes which encode the projections of the underlying stress tensor field.}
    \label{fig:render_multiple_views}
\end{figure}

In Sec.~\ref{sec:integrated_photoelasticity}, we showed that the equivalent Jones matrix (Eq.~\ref{eq:J_eq})   for each ray encodes a projection of the stress distribution. Here we describe how to measure this Jones matrix through a multi-axis polariscope capture setup comprising a linear polarizer and quarter waveplate before and after the target. We then describe how rotating the object and capturing polariscope measurement for each rotation enables photoelastic tomography.

\paragraph{Circular polariscope}
We consider a circular polariscope-based setup (Fig.~\ref{fig:render_multiple_views}) to capture components of the Jones matrix in Eq.~\ref{eq:J_eq}. Light emitted by an area light source passes through a linear polarizer (LP1) with the polarization axis oriented at an angle $\alpha_1$ with respect to the horizontal axis. The linearly polarized light then passes through a quarter wave plate (QWP1) with the fast axis oriented at an angle $\beta_1$ with relative to the horizontal axis. The resulting circularly polarized light passes the specimen under stress. The light ray coming out of the specimen passes through another quarter wave plate (QWP2) with fast axis oriented at an angle $\beta_2$ and then reaches the polarization camera. The polarization camera comprises of a grid of linear polarizers at different orientations. Consider this light ray hits a sensor pixel which has a linear polarizer (LP2) oriented at an angle $\alpha_2$.

\paragraph{Captured intensity}
Consider $\mathbf{E}(\alpha_1)$ as the Jones vector of light emitted from the LP1 and $\mathbf{Q}(\beta_1)$,$\mathbf{Q}(\beta_2)$ and $\mathbf{L}(\alpha_2)$ as the Jones matrices of QWP1, QWP2 and LP2 respectively. The effective Jones matrix as derived in Eq.~\ref{eq:J_eq} depends on the characteristic parameters $\delta_{net}$, $\theta_{net}$ and $\gamma_{net}$.
The refraction into and outside the object also alters the polarization state based on the surface normals at those points. At the entry and exit points of the ray, $t_n$ and $t_f$, consider the Jones matrix depending on the surface normals $\mathbf{R}(t_n)$  and $\mathbf{R}(t_f)$.
Using Jones Calculus, the Jones vector of light reaching the sensor $\mathbf{E}_\text{pol}$ is given by the complex matrix multiplication:
\begin{align}
    \mathbf{E}_{\text{pol}} = \mathbf{L}(\alpha_1)\mathbf{Q}(\beta_1)\mathbf{R}(t_n)\mathbf{J}_{\text{eq}}\mathbf{R}(t_f)\mathbf{Q}(\beta_2)\mathbf{E}(\alpha_2)
    \label{eq:E_cap}
\end{align}
The intensity measurement captured by the polariscope $I_\text{pol}$ can be obtained from the Jones vector $\mathbf{E}$ by combining net intensity along orthogonal dimensions $\mathbf{u}$ and $\mathbf{v}$. 
\begin{align}
    I_\text{pol} = E_\mathbf{u}E_\mathbf{u}^* + E_\mathbf{v}E_\mathbf{v}^*
    \label{eq:I_cap}
\end{align}
By varying the polariscope element angles $\alpha_1$, $\beta_1$, $\beta_2$ and $\alpha_2$, we can obtain multiple intensity measurements that encode the characteristic parameters $\delta_{\text{eq}}$, $\theta_{\text{eq}}$ and $\gamma_{\text{eq}}$.

\paragraph{Multi-axis rotations}
The polariscope measurements encode the equivalent Jones matrix $\mathbf{J}_\text{eq}$ (Eq~\ref{eq:J_eq}) for each camera ray. This Jones matrix in turn depends on the projection of the stress tensors of the points along the ray (Eq~\ref{eq:projected_stress}). Every ray measures only a $2\times2$ projection of the $3\times3$ stress tensors on a plane perpendicular to that ray (Eq.~\ref{eq:projected_stress}). Therefore, to obtain the full $3\times3$ stress tensor at each point, we need to probe every point with multiple rays, each providing a different $2\times2$ projection of the $3\times3$ tensor. To obtain all possible $2\times2$ projections, the probing rays should be uniformly sampled from a unit hemisphere. 

To obtain multiple projections of the stress tensor field, we could either rotate the viewing camera rays by moving the camera and the screen around the object or we could rotate the object. In Fig.~\ref{fig:render_multiple_views}, we visualize the later case. Yaw rotation $\rho_1$ and pitch rotation $\rho_2$ of the object are equivalent to fixing the object and rotating the elevation and azimuth direction of the camera ray respectively. Thus by multiple yaw-pitch rotations of the object, we capture multiple projections of the stress tensor field enabling photoelastic tomography. 

\paragraph{Summarizing multi-axis polariscopy}
Putting it all together, our capture scheme (Fig.~\ref{fig:render_multiple_views}) has two components: 1) yaw-pitch rotations of the object $\rho_1$, $\rho_2$  and 2) polariscope images involving rotations of LP1($\alpha_1$), QWP1 ($\beta_1$), QWP2 ($\beta_2$) and LP2 ($\alpha_2$) for each yaw-pitch rotation. Each captured intensity is a non-linear function $\mathcal{F}$ of the characteristic parameters $\delta_\mathrm{eq},\theta_\mathrm{eq},\gamma_\mathrm{eq}$ (Eq.~\ref{eq:E_cap}-\ref{eq:I_cap}). These characteristic parameters in turn depend on the underlying stress tensor distribution $\mathbf{S}$ (Eq.~\ref{eq:projected_stress}-\ref{eq:J_eq}). We can express the captured polariscope measurements as :
\begin{align}
   I_\text{pol}\left(\bm{\alpha},\bm{\beta}, \bm{\rho} \right) = \mathcal{F}\left(\mathbf{J}_\text{eq}\left(\delta_\text{eq}(\mathbf{S}), \theta_\text{eq}(\mathbf{S}), \gamma_\text{eq}(\mathbf{S})\right) \right) \; ,
   \label{eq:I_pol_F}
\end{align}
where we have folded the polariscope parameters into vectors,
\begin{align}
    \bm{\alpha} = [\alpha_1, \alpha_2] \quad \bm{\beta} = [\beta_1, \beta_2] \quad \bm{\rho} = [\rho_1, \rho_2] \; .
\end{align}
The characteristic parameters Fig.~\ref{fig:render_multiple_views}(b) shows example rendered polariscope images from our capture scheme for a disk under diametral compression.

\section{Neural Stress Tensor Tomography}
\begin{figure}
    \centering
    \includegraphics{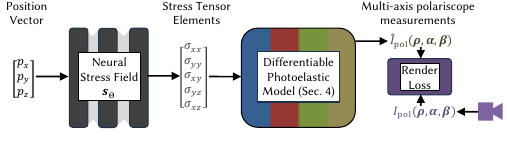}
    \caption{\textbf{NeST Pipeline.} For every sampled point, the 3D position vector is passed through a coordinate-based neural network, called the neural stress field, to obtain the stress tensor elements at that point. Our differentiable forward model converts the queried stress to multi-axis polariscope measurements that we compare with the captured measurements and use the render losss to end-to-end train the underlying neural stress field. }
    \label{fig:our_pipeline}
\end{figure}

Here we describe our approach to reconstruct the 3 dimensional stress tensor field from multi-axis polariscope measurements. We first model the stress tensor field $S(\mathbf{p})$ as a neural implicit representation and describe how we can use the developed 3D photoelasticity model to render the projected Jones matrix $\mathbf{J}_{\text{net}}$ of the sample from these representations and then show how we can solve for unknown neural stress fields with gradient-based optimization. 
\subsection{Neural stress fields}
 The stress field can vary dramatically throughout the object. For example, when an external load is applied on the object's boundary stress is often concentrated close to the boundary and then becomes sparse towards the bulk of the object (Fig.~\ref{fig:cir_pol}). Representing this stress distribution requires adaptive sampling points in the object. For the stress tomography problem, as the stress is completely unknown at the start, this adaptive sampling cannot be fixed and known. We leverage the recent advancements in implicit neural representations to model complex visual distributions with high expressive power and computational efficiency.
 
\begin{figure*}
    \centering
    \includegraphics{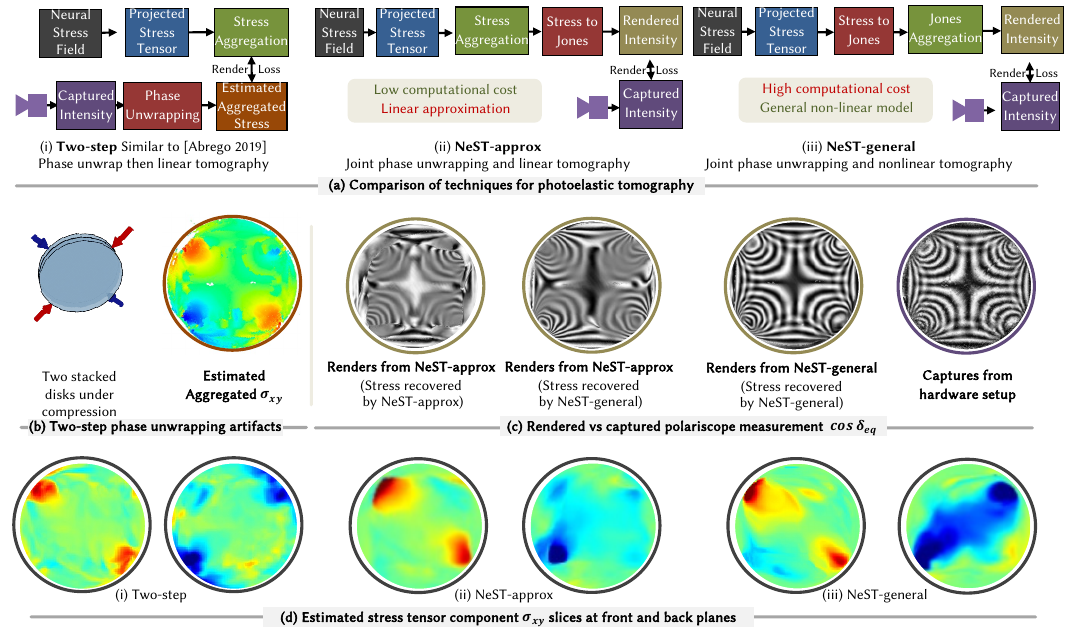}
    \caption{\textbf{Comparison of approaches for photoelastic tomography.} The existing approaches to photoelastic tomography \cite{abrego2019experimental} (a(i)) involve a separate per-view phase unwrapping estimation that suffers from artefacts (b). We propose a joint phase unwrapping and tomography approach. Leveraging our linear forward model (a(ii)) results in more approximate renders while being computationally efficient. Our general forward model (a(ii)) results in renders that closely match the captures (c) at a higher computational cost. Moreover the estimated stress field with the general model is smoother compared to the other approaches (d).}
    \label{fig:pipeline}
\end{figure*}
 Stress at each point $\mathbf{p}$ within the object is modeled as a second-order symmetric tensor (Eq.~\ref{eq:stress_tensor}). We express this distribution with a coordinate-based MLP network $S$ with weights $\Theta$ that takes the position $\mathbf{p}$ as input and outputs five components of the Stress tensor matrix :
 \begin{align}
    S_{\Theta} : \mathbf{p} \rightarrow (\sigma_{xx}, \sigma_{yy}, \sigma_{xy}, \sigma_{yz}, \sigma_{zx}) \;.
 \end{align}
\paragraph{Handling trace ambiguity} The stress tensor estimated by photoelasticity has an unknown offset to the trace of the stress tensor matrix \cite{lionheart2009reconstruction}. This is evident from the stress optic equation (Eq.~\ref{eq:stress_optic}) where the phase difference depends on the difference between principal stresses. The addition of a constant offset to the principal stress would still result in the same phase difference. This ambiguity could result in many plausible stress tensors to fit the measurements. We account for this ambiguity by explicitly setting the trace of the reconstructed stress tensor as zero. As a result, we estimate the sixth component of the stress tensor $\sigma_{zz}$ so that the trace, $\sigma_{xx} + \sigma_{yy} + \sigma_{zz} $, is zero. 
\begin{align}
    \sigma_{zz} = -\sigma_{xx} - \sigma_{yy} \;.
\end{align}
\paragraph{Occupancy function}   
We consider that the 3D geometry of the loaded specimen is known and represented as a known occupancy function $O$ that is 0 for any point outside the object and 1 for any point on and within the object. The queried stress at a point is then masked with this occupancy function. The occupancy function aids the reconstruction of neural stress field by explicitly setting the stress at empty regions to zero. We denote the masked stress field as $S^\text{m}_{\Theta}$:
\begin{align}
    S^\text{m}_{\Theta}(\mathbf{p})= O(\mathbf{p})S_{\Theta}(\mathbf{p})
\end{align}

\begin{figure*}
    \centering
    \includegraphics[width=\textwidth]{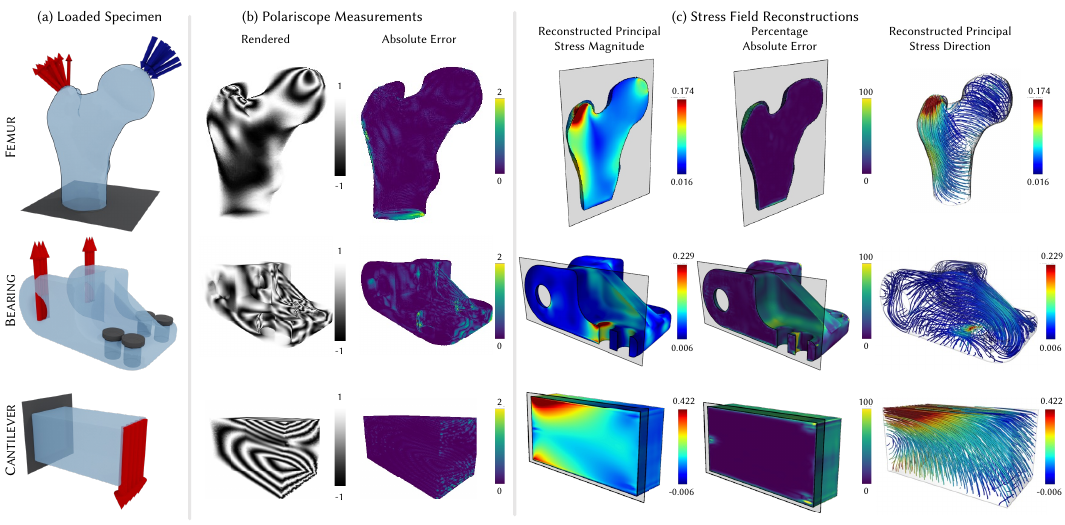}
    \caption{\textbf{NeST evaluation on simulated dataset.} We evaluate NeST on multi-axis polariscope measurements rendered from stress fields in common objects with varying loading conditions (a). The measurements rendered from the stress field estimated by NeST qualitatively match the input measurements (b). In (c), we visualize the major principal component of the stress tensor field estimated by NeST by showing a slice of the principal stress magnitude and streamline visualization \cite{wang20223d} of the principal stress direction. (b) and (c) demonstrate that the stress field estimated by NeST qualitatively and quantitatively matches the underlying ground truth stress field. Please refer to the supplementary video for a 3D animation of this result.}
    \label{fig:sim_showcase}
\end{figure*}
\subsection{Differentiable rendering of polariscope measurements from neural stress fields}
\label{sec:diff_rendering_forward_model}
Our differentiable formulation for 3D photoelasticity (Sec.~\ref{sec:integrated_photoelasticity}) is well-suited for rendering from continuous implicit stress distributions. Here we describe how we utilize the differentiable forward model in our optimization framework. 
\paragraph{Monte-Carlo ray sampling}
We require to obtain the polariscope measurements for each multi-axis rotation  $\bm{\rho}$ of the object. As described in Sec.~\ref{sec:polariscope}, the measurements can equivalently be modeled by fixing the object and rotating the polariscope assembly by $-\bm{\rho}$. Fixing the object allows us to query the neural stress field $S^\text{m}_{\Theta}$ in the same coordinate system for all the rotations. For each rotation $\bm{\rho}$, we obtain a set of rays corresponding to each pixel on the camera and we parameterize the ray as $\bm{o},\bm{d},\bm{\rho}$.

For each ray $(\mathbf{o}, \mathbf{d}, \bm{\rho})$ we sample $N$ points using stratified sampling between the object boundaries $(t_n, t_f)$. For every point $t$, we query the stress tensor $S^\text{m}_{\Theta}(t)$ and obtain the Jones matrix $\mathbf{J}(t)$ from Eq.~\ref{eq:J_retarder}.
From Eq.~\ref{eq:integrated_photoelasticity_Jones}, the projected Jones matrix can be obtained by complex matrix multiplications of Jones matrices along the ray
\begin{align}
    \mathbf{J}_\text{eq} = \prod_{i=1}^{N} \mathbf{J}(t_i) \;.
    \label{eq:J_net_i}
\end{align}

\paragraph{Efficient Jones matrix multiplications}
Multiplying complex $2\times2$ Jones matrices in Eq.~\ref{eq:J_net_i} along the ray can be computationally expensive. Multiplying two Jones matrices itself requires 56 multiply-add operations. We show that these computations can be simplified by using the Poincaré theorem and exploiting the structure of these Jones matrices. 
The projected Jones matrix $\mathbf{J}_\text{net}$ (Eq.~\ref{eq:J_eq}) has only 4 unique scalar elements, $\mathbf{j}_\text{eq} = (a_\text{eq},b_\text{eq},c_\text{eq},d_\text{eq})$, defined as
\begin{align}
    a_\text{eq} &= \cos{\left(\nicefrac{\delta_\text{eq}}{2}\right)}\cos 2\gamma_\text{eq}   & 
    b_\text{eq} &= \sin{\left(\nicefrac{\delta_\text{eq}}{2}\right)}\sin 2\theta'_\text{eq}  \notag \\
    c_\text{eq} &= \sin{\left(\nicefrac{\delta_\text{eq}}{2}\right)}\cos 2\theta'_\text{eq}  &
    d_\text{eq} &= \cos{\left(\nicefrac{\delta_\text{eq}}{2}\right)}\sin 2\gamma_\text{eq}
    \label{eq:abcd_eq}
\end{align}
Similarly, for each intermediate Jones matrix $\mathbf{J}(t_i)$ from Eq.~\ref{eq:J_retarder} can be defined with unique elements  $ \mathbf{j}_i = (a_i, b_i, c_i)$. 
\begin{align}
    a_i &= \cos{\left(\nicefrac{\delta_i}{2}\right)}   & 
    b_i &= \sin{\left(\nicefrac{\delta_i}{2}\right)}\sin 2\theta_i  \notag \\
    c_i &= \sin{\left(\nicefrac{\delta_i}{2}\right)}\cos 2\theta_i  &
    \label{eq:abcd_i}
\end{align}
Consider the first $i$ points along the ray. From the Poincaré theorem, the Jones matrices for these points are equivalent to a single equivalent Jones matrix that we denote with unique elements $\mathbf{j}^{i+1}_\text{eq} = (a^i_\text{eq},b^i_\text{eq},c^i_\text{eq},d^i_\text{eq})$. The equivalent Jones matrix for $i+1$ points can be obtained by matrix multiplication of equivalent Jones matrix upto $i$ points with Jones matrix $\mathbf{J}(t_i)$. As before, the equivalent matrix for $i+1$ points is represented its by unique elements $\mathbf{j}^{i+1}_\text{eq}$ and the matrix multiplication can be efficiently expressed as a function the unique elements upto $i$ points and the unique elements of the $i$th point:
\begin{align}
    a^{i+1}_\text{eq} &= a_ia^{i}_\text{eq} - b_ib^{i}_\text{eq} - c_ic^{i}_\text{eq} \\
    b^{i+1}_\text{eq} &= a_ib^{i}_\text{eq} + b_ia^{i}_\text{eq} + c_id^{i}_\text{eq} \\
    c^{i+1}_\text{eq} &= a_ic^{i}_\text{eq} + c_ia^{i}_\text{eq} - b_id^{i}_\text{eq} \\
    d^{i+1}_\text{eq} &= a_id^{i}_\text{eq} - c_ib^{i}_\text{eq} + b_ic^{i}_\text{eq} 
\end{align}
Accumulating these unique elements up to $N$ points, we obtain the unique elements of the overall equivalent Jones matrix,$\mathbf{j}_\text{eq} = \mathbf{j}^N_\text{eq}$. This approach results in $~ 3\times$ reduction of the total number of computations compared to naive complex matrix multiplications. 

\begin{figure*}[t]
    \centering
    \includegraphics[width=\linewidth]{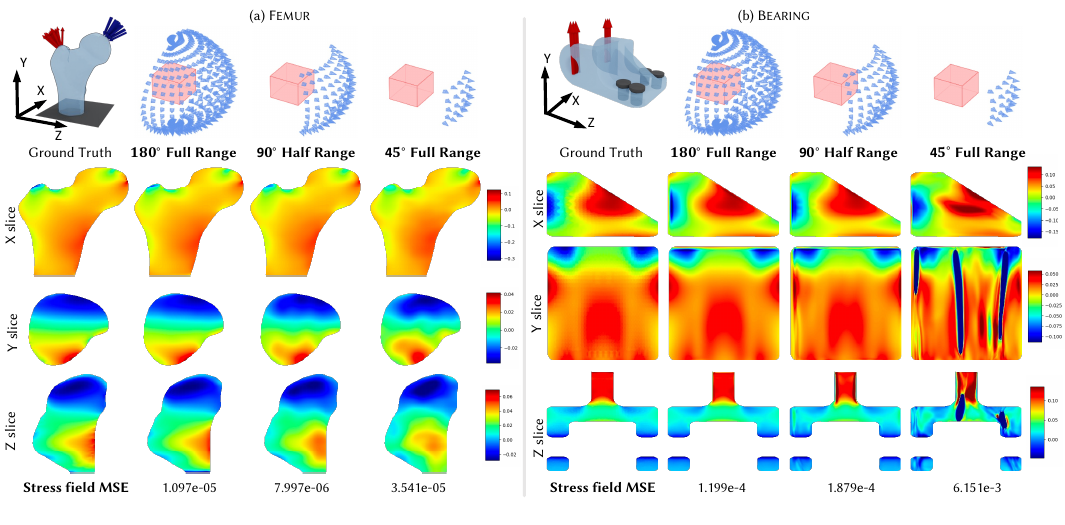}
    \caption{\textbf{The effect of varying rotation angles.} We vary the the range of multi-axis rotation angles for two scenes in our simulated dataset and show slices of the $\sigma_xx$ component. A smaller range results in lower reconstruction accuracy. $\textsc{Bearing}$ has more complex stress variation than $\textsc{Femur}$ and thus detoriates higher upon reducing the rotation angle range. }
    \label{fig:sim_analysis_rotation}
\end{figure*}
\paragraph{Multi-axis polariscope measurements}
For the ray $(\bm{o},\bm{d},\bm{\rho})$, we then compute rendered polariscope measurements $\hat{I}_\text{pol}(\bm{\alpha}, \bm{\beta}, \bm{\rho})$ from the computed equivalent Jones matrix parameterized with the vector $\mathbf{j}_\text{eq}$ using Eq.~\ref{eq:I_pol_F} as  
\begin{align}
    \hat{I}_\text{pol}(\bm{\alpha}, \bm{\beta}, \bm{\rho})  =  \mathcal{F}\left(\bm{j}_\text{eq}\left( \bm{S}_\Theta\right)\right).
\end{align}

\subsection{Optimization objective}
 From our capture setup, we obtain the captured multi-axis polariscope $I_{\text{pol}}$ for every rotation $\bm{\rho}$ and polariscope parameter $\bm{\alpha},\bm{\beta}$.
We define the loss between the rendered polariscope measurements $\hat{I}_{\text{pol}}$ from Sec.~\ref{sec:diff_rendering_forward_model} and the captured $I_{\text{pol}}$ as the L1 loss and optimize for the parameters of the neural stress field $\Theta$ using gradient-based optimization. The estimated parameters $\Theta^*$ can be expressed as:
\begin{align}
    \Theta^*= \min_{\Theta}  \sum_{\bm{\alpha},\bm{\beta},\bm{\rho}} \| \hat{I}_{\text{pol}}(\Theta) - I_{\text{pol}} \|_1.
\end{align}

\subsection{Comparison of photoelastic tomography approaches}
Here we compare and contrast three different approaches to photoelastic tomography enabled by our optimization framework (Fig.~\ref{fig:pipeline}(a)). We use the example of experimental data that consists of two planar disks that are under diametral compression and are stacked on top of each other. When slicing along the thickness of the object underlying stress distribution should rotate from $135^\circ$ to $45^\circ$ as we go from one planar disk to the other. We compare the following three techniques:
\begin{itemize}[leftmargin=*]
    \item \texttt{Two-step}: Similar to the prior linear tensor tomography technique by Abrego et al. \shortcite{abrego2019experimental}, we separately phase unwrap polariscope measurements for each rotation $\bm{\rho}$ and obtain the aggregated stress. These aggregated stresses are then used to solve a linear tensor tomography problem using our linear forward model and neural stress fields.
    \item \texttt{NeST-approx}: We use our differentiable linear forward model and perform joint phase unwrapping and linear tomography.   
    \item \texttt{NeST-general}: We use the general nonlinear forward model and perform joint phase unwrapping and nonlinear tomography.
\end{itemize}

\texttt{Two-step} has artifacts in the aggregated stress obtained after phase unwrapping (Fig.~\ref{fig:pipeline}(b)). While \texttt{NeST-approx} is more computationally efficient by using the linear model instead of the nonlinear one, it cannot explain the captured measurements as well as the nonlinear model Fig.~\ref{fig:pipeline}(c)). \texttt{NeST-general} can accurately explain the captured measurements and reconstruct stress fields that are smoother and qualitatively closer to the expected variation (Fig.~\ref{fig:pipeline}(d)) than other two techniques.
\begin{figure*}
    \centering
    \includegraphics[width=\textwidth]{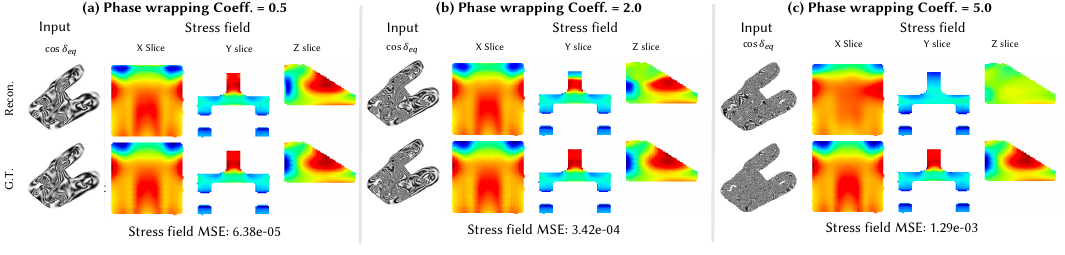}
    \caption{\textbf{The effect of increasing phase wrapping.} More fringes can be observed in the rendered image as the wraping coefficient increases. The reconstruction quality detoriates when the fringes become too dense for the a given spatial resolution. }
    \label{fig:sim_analysis_coeff}
\end{figure*}

\subsection{NeST Implementation Details}
\paragraph{Optimization framework}
We implement the NeST framework in PyTorch. We use NerfAcc \cite{li2023nerfacc} as the backbone of our implementation. NerfAcc accelerates NeRF-based reconstruction using CUDA kernels. Our forward model involves unique cascaded complex multiplications that we perform efficiently using the procedure defined in Sec.~\ref{sec:diff_rendering_forward_model}. We implement the forward and gradient operators for our image formation model using custom CUDA kernels.
\paragraph{Neural field architecture}
We consider the neural stress fields as coordinate-based MLP with sinusoidal encodings \cite{mildenhall2021nerf} and sigmoid linear unit (SiLU) activation function \cite{hendrycks2016gaussian}. For simulated experiments, we consider an 8-layer MLP with 64 neurons in each layer and 5 frequencies in the sinusoidal encoding. For the real experiments, we consider a 6-layer MLP with 64 neurons and 4 sinusoidal encoding frequencies. 
\paragraph{Occupancy grid}
Similar to InstantNGP \cite{muller2022instant}, we define a binary occupancy grid that ensures that the points on rays that are empty are not queried. InstantNGP learns the occupancy grid jointly with the neural field. In our case, as the geometry is known, we define this occupancy grid from on the occupancy function $O$.
\paragraph{Reconstruction details}
For optimization we use an Adam optimizer, with a learning rate of 3e-4. We optimize for about 100K iterations per scene. Optimizing each scene takes around 2 hours on an NVIDIA A100 GPU.   


\section{Simulation Results and Analysis}
In this section, we discuss the evaluation of NeST on synthetic datasets. We describe the generation of our 3D photoelasticity dataset from stress fields of complex objects, evaluate the performance of NeST on this dataset, and analyze the reconstruction accuracy under key factors.
\subsection{Simulated 3D Photoelasticity Dataset}




\paragraph{3D-TSV stress fields}
We utilize the 3D-TSV dataset \cite{wang20223d} that includes 3D stress fields of several common objects and mechanical parts generated under practical loading conditions using finite element method (FEM) simulations. The stress field within each object is represented by an adaptively sampled hexahedral mesh or a Cartesian grid. The six-element symmetric stress tensor elements are provided at the vertices of the adaptively sampled mesh. The dataset also contains the 3D surface mesh and the details of the loading conditions for each object. We use six objects from the 3D-TSV dataset to construct our 3D TSV dataset: \textsc{Femur}, ~\textsc{Bearing},~\textsc{Cantilever},~\textsc{Kitten},~\textsc{Bracket} and \textsc{Rod}. The first three datasets are depicted in Fig.~\ref{fig:sim_showcase} while the others are depicted in the supplement. 

\paragraph{KNN interpolation}
The stress field in 3D TSV datasets is defined only on 3D coordinates corresponding to the vertices of an adaptively sampled mesh. Our rendering procedure described in Sec.~\ref{sec:image_formation_model} requires us to query the stress field at 3D arbitrary points within the mesh. We use k-Nearest Neighbors (KNN) interpolation \cite{qi2017pointnet++} to compute the stress tensor elements at any arbitrary interior point by distance-based interpolation of the stress values at k nearest vertices. 

\paragraph{SIREN occupancy function}
The KNN interpolation assigns non-zero stress values to 3D points outside the object. We require an occupancy function to set the stress values outside the object to 0 explicitly. First, we train a SIREN network \cite{sitzmann2020implicit} to model the signed distance function (SDF) from the provided surface mesh. Then we obtain the occupancy function by thresholding the SDF function such that the non-negative values correspond to 1 and the positive values correspond to 0. We use a SIREN network with three hidden layers and 256 neurons for all the objects. This occupancy function is used in both (1) the rendering stage to mask stress field values directly and (2) the reconstruction stage to exclude low occupancy regions and accelerate sampling.

\paragraph{Rendering 3D photoelasticity}
We query the stress field at arbitrary points with the procedure described above. We use the nonlinear forward model described in Sec~\ref{sec:diff_rendering_forward_model} to render multi-axis polariscopy measurements. We can vary the stress-optic coefficient $C$ (Eq.~\ref{eq:stress_optic}) to vary the frequency of photoelastic fringes.  Fig~\ref{fig:cir_pol}(c) shows rendered measurements for \textsc{Femur}, with the underlying stress field depicted in Fig~\ref{fig:cir_pol}(a). As expected, the photoelastic fringes in the rendered measurements have a higher density around the load application points. The multi-axis polariscope measurements for each object are rendered for the complete 180-degree range for the azimuth and elevation rotation axis with 32 angles sampled along each rotation axis. Rendering each object takes 2-4 hours on an Nvidia A100 GPU. 



\subsection{Evaluation of reconstruction}

We use the $32\times32$ renderings with the full 180-degree range on each axis for qualitative comparison. The reconstructions are performed using our \texttt{NeST-general} approach. In Fig. \ref{fig:sim_showcase}(a), we show the object geometry and loading conditions for three objects:~\textsc{Femur},~\textsc{Bracket} and \textsc{Cantilever}. Blue arrows represent compressive forces and red arrows depict tensile and shear forces. The object points touching the gray surfaces are kept fixed during the load application. Fig.~\ref{fig:sim_showcase}(b) shows that the polariscope measurements rendered from the reconstructed stress field by NeST qualitatively match those rendered directly from the ground truth field. In Fig.~\ref{fig:sim_showcase}(c), we visualize the reconstructed stress plotting the magnitude and directions of the principal stress corresponding to the largest eigenvalue of the stress tensor and its eigenvector respectively. We can see that our reconstruction closely approximates the ground truth. From the principal stress magnitude, we can observe the increase in stress near the contact points on \textsc{Femur}, around the holes in \textsc{Bracket}, and at the top and bottom edges in \textsc{Cantilever}. The principal stress direction is visualized as stress lines using the 3D-TSV visualization framework \cite{wang20223d} and demonstrates how the stress propagates within the object.

\subsection{Effect of rotation angles}
In tomography, the rotation angle or the scanning angle is important. Here, we start with the hemisphere (180$\times$180), and gradually reduce to a cone where the center aligns with the object. We reduce the angle of the cone (scanning range in each direction) from 180 down to 90 degrees on each direction (2$\times$ subsampling), and 45 degrees (4$\times$ subsampling). The effects are shown in Fig. \ref{fig:sim_analysis_rotation} below.  As the angle reduces, the quality of reconstruction reduces significantly. With 90 degrees, reasonable reconstruction can be obtained for most of the object. However, artifacts start to occur. When subsampling is 4$\times$, the quality degrades significantly and distortions can be seen in the reconstructed stress field. The distortions in \textsc{Bearing} are much more severe as it has a more complex stress field. 
The results match the observations in conventional scalar tomography, where the missing cone problem can significantly degrade the performance as unobserved angles increase.

\subsection{Effect of phase wrapping}
We analyze the effect of increasing phase wrapping using \textsc{Bearing}. We vary the stress-optic coefficient for \textsc{Bearing} in the rendering stage from $0.1$ to $0.25$ to $0.5$. To emulate sensor read noise in real captures, we add a Gaussian noise with mean $0$ and standard deviation $0.01$. Example ground truth rendered measurements and slices of the underlying stress field are shown in the first row of Fig.~\ref{fig:sim_analysis_coeff}. Increasing the stress-optic coefficient increases the amount of phase wrapping in the reconstructions. For a given spatial resolution, at a certain coefficient value, the fringes become too dense to be resolved resulting in detoriation of the reconstruction quality. 

\section{Real Experiments and Results}

\begin{figure}
    \centering
    \includegraphics[width=\linewidth]{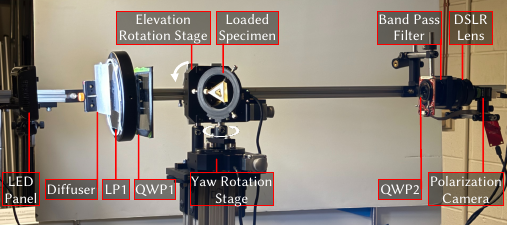}
    \caption{\textbf{Proposed multi-axis polariscope setup.} The system is composed of a linearly polarized light panel for illumination and polarization camera for detection. Both illumination and detection side additionally include rotatable quarter wave plates. The entire assembly is colinearly mounted. Two-axis object rotation is achieved via a motorized azimuth stage ($\rho_2$) and by rotating the entire polariscope assembly for elevation ($\rho_1$).
    }
    \label{fig:acquisition_setup}
\end{figure}
\begin{figure}
    \centering
    \includegraphics{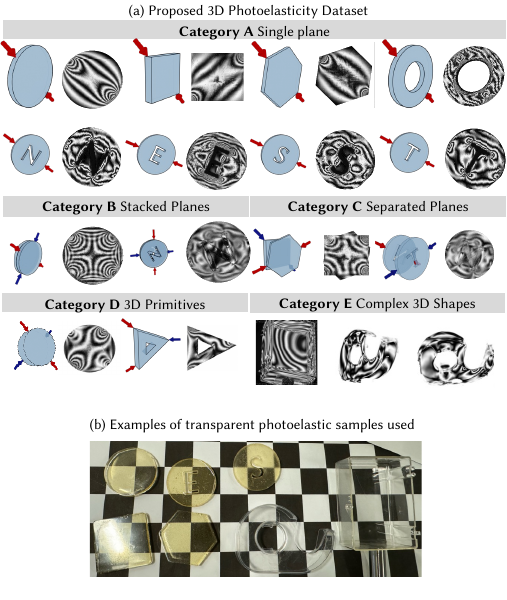}
    \caption{\textbf{Proposed 3D Photoelasticity experimental dataset.} We acquire multi-axis polariscope measurements for a variety of 3D shapes and force application conditions. (a) depicts examples from the dataset with the measurement $\cos{\delta_\text{eq}}$. The photoelastic test specimen are created from epoxy resin and shown in (b). }
    \label{fig:real_scenes}
\end{figure}
\begin{figure*}
    \centering
    \includegraphics[width=\linewidth]{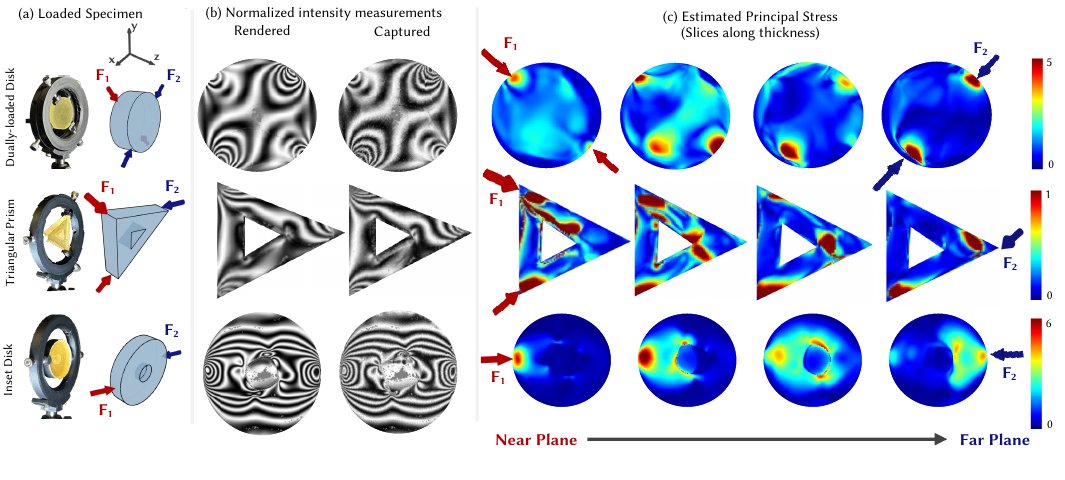}
    \caption{\textbf{NeST evaluation on real experiments.} We demonstrate that NeST can recover the internal stress field distribution for 3D objects with varying shapes and loading conditions (a). Polariscope measurement $\cos \delta_\text{eq}$ rendered from the stress field estimated by NeST matches the input measurements (c). Slices of the principal stress distribution qualitatively match the intuition for the applied loading conditions. Please refer to the supplementary video for a 3D animation of this result.}
    \label{fig:real_qualitative_disk}
\end{figure*}

\subsection{Acquisition setup}

Our multi-axis polariscope acquisition setup is shown in Fig.~\ref{fig:acquisition_setup}. On the illumination side, our polariscope assembly has an LED light panel with a diffuser, a linear polarizer (LP1), and a quarter-wave plate (QWP1). On the camera side, our setup has another quarter-wave plate (QWP2), a bandpass filter, DSLR lens, and a snapshot polarization sensor (containing a grid of LP2). The LP1 and QWP1 can be manually rotated while QWP2 is rotated with a motorized stage. The entire polariscope assembly is mounted colinearly on a single rod.

As explained in Sec.~\ref{sec:polariscope}, we require polariscope measurements by rotating the object along two axes: azimuth $(\rho_2)$ and elevation ($\rho_1$). We rotate the object along the azimuth axis with a motorized rotation stage. For the elevation axis, we rotate the entire polariscope assembly with second motorized rotation stage. We rotate the whole assembly instead of just the object to minimize the spatial footprint of the setup and avoid occluding the object with the stage. 

We capture four raw measurements for each elevation and azimuth rotation with varying QWP1 and QWP2 orientations. Each of the four raw measurements comprises four different LP2 orientations. These 16 measurements $I_\text{pol}$ are non-linear expressions of the underlying characteristic parameters $\delta_\text{eq}$, $\theta_\text{eq}$ and $\gamma_\text{eq}$ (Eq.~\ref{eq:I_pol_F}). Please refer to the supplement document for the analytical forms of all 16 measurements. As we describe in the supplement, these measurements can be expressed as constant scale and offset applied to one of the following six expressions of the characteristic parameters, $\tilde{I}_i$:
\begin{align}
   \tilde{I}_1 &= \cos \delta_\text{eq} & 
   \tilde{I}_2 = \sin \delta_\text{eq} \cos 2\theta_\text{eq} \\
   \tilde{I}_3 &= \sin \delta_\text{eq} \sin 2\theta_\text{eq} & \tilde{I}_4 = \sin \delta_\text{eq} \sin 2\gamma'_\text{eq} 
\end{align}
\begin{align}
   \tilde{I}_5 &= \cos 2\theta_\text{eq} \sin 2\gamma'_\text{eq} + \cos \delta_\text{eq} \sin 2\theta_\text{eq} \sin 2\gamma'_\text{eq}  \\
   \tilde{I}_6 &= \sin 2\theta_\text{eq} \cos 2\gamma'_\text{eq} - \cos \delta_\text{eq} \cos 2\theta_\text{eq} \sin 2\gamma'_\text{eq}   \;,
\end{align}
where $\gamma'_\text{eq} = 2\gamma_\text{eq} - \theta_\text{eq}$.


\subsection{Test samples}

The custom test samples were made from a two-part epoxy resin and molded in various shapes using silicon molds. For simple shapes, off-the-shelf molds were utilized. For more complex shapes, including samples with letters or with strict dimensions, liquid silicon molds were cast around 3D printed parts, and the samples were then created in the same way as before. 

\paragraph{Samples with Residual Stresses}
Two types of samples with residual stresses were used: household transparent plastics and resin. The household objects were simply objects that are commonly found around the house, such as a tape dispenser, and transparent plastic box. Due to the injection molded process used to manufacture these parts, residual stresses are common and the samples with the most birefringence were selected. The second type was resin samples. These were cast resin samples that had a compressive force applied for in excess of 24 hours. Due to the nature of the resin we used, the samples would set and exhibit residual stresses even after the force was removed. These samples simplified the mounting 
setup and allowed for extended azimuth and elevation ranges. 

\paragraph{Samples with Applied Stresses}

To apply an adequate force, custom 3D-printed mounts were made which applied a compressive force on the samples. The mount was modeled such that it allows varying angles of applied stress on the XY-plane. Furthermore, the mount was created such that two samples could be mounted sequentially along Z, as shown in Fig~\ref{fig:acquisition_setup}. Each force applicator can be removed independent of the other to facilitate the capture of individual samples for validation: the front sample was captured, both samples were then captured together, and then the back sample was captured. To restrict the force applied to the sample to compression, a ball-and-socket nut was mounted to each screw such that no torsion was applied as the screw was tightened.  
\begin{figure*}
    \centering
    \includegraphics{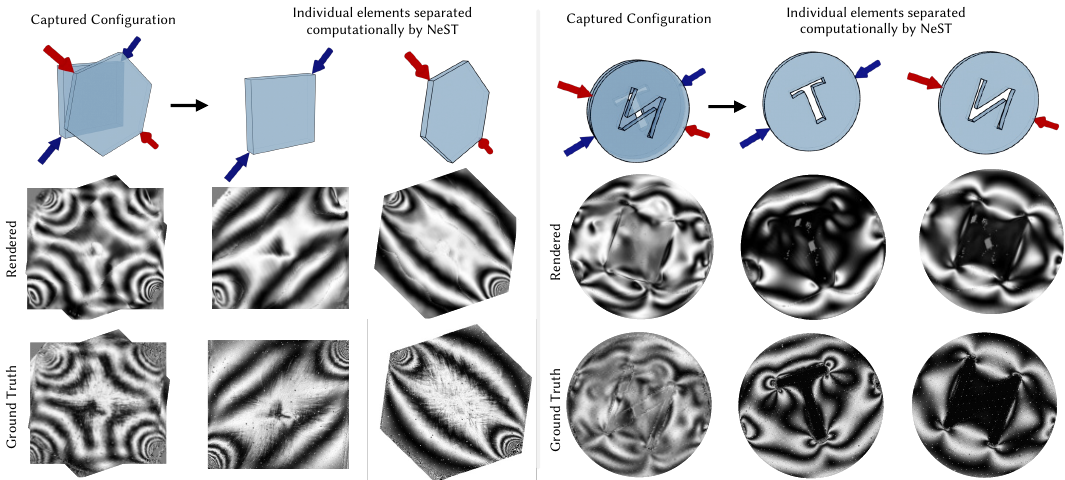}
    \caption{\textbf{NeST Application: Virtual slicing.} NeST enables visualizing the internal stress distribution as photoelastic fringes obtained by virtually slicing the object. For objects consisting of two stacked compressed planar elements, NeST can decompose the stress field for each planar element and recover the photoelastic fringes for each element. These decomposed photoelastic fringes match the ones captured by scanning each element separately.  }
    \label{fig:app_virtual_slicing}
\end{figure*}

\subsection{Acquisition Details}
For the samples with residual stress, we capture the 16 polariscope measurements for each of the 400 multi-axis rotations: 25 azimuth and 16 elevation rotations with a range of 180 and 90 degrees respectively.
320 rotations are used for training and 80 rotations for testing. For the samples with applied stress, the force application mount occludes the sample at oblique views and thus limits the azimuth range. These samples are captured with 320 multi-axis rotations: 20 azimuth and 16 elevation rotations with a range of 140 and 90 degrees respectively. The entire acquisition requires around 3 hours for each object.

\subsection{Qualitative Results}

In Fig.~\ref{fig:real_qualitative_disk}, we evaluate the performance of NeST in recovering underlying 3D stress distributions in real specimens under load. We analyze three loaded specimens (a): 1) a cylinder with two compressive loads along the thickness close to each face and 2) a triangular prism with loads partially along z (i.e. a shear load in addition to a compressive load is applied).  3) a cylinder with a square hole and with a single obliquely applied compressive load. All these specimens correspond to a 3D stress variation as the load application points are varying along the thickness. We reconstruct the underlying stress tensor field using nest. The polariscope measurement from the reconstructed stress field matches the input capture (b). We compute the principal stress field from the recovered stress tensor and plot its slices along the thickness of the specimen (c). From these slices, is it evident that NeST can resolve the stress fields corresponding to different load points with recovered principal stress being higher close to the respective load points, depicted as arrows in (c). 

\subsection{NeST Applications}

\paragraph{Virtual Slicing}
NeST enables virtually slicing 3D objects to visualize the internal stress within the object as photoelastic fringes. An example of virtual slicing using NeST is shown in  Fig.~\ref{fig:app_virtual_slicing}. With two objects stacked in line with one on top of another, the photoelastic fringes are superimposed on one another and interfere. NeST is able to successfully decompose the fringes of each specimen, and their individual components can be extracted. This is shown with a simple scenario with compressed square and hexagonal prisms. A more complex example was tested where individual letter cutouts were extracted from specimens, which is challenging given the non-uniformity of the profiles around the sharp edges of the letters.

\begin{figure*}
    \centering
    \includegraphics{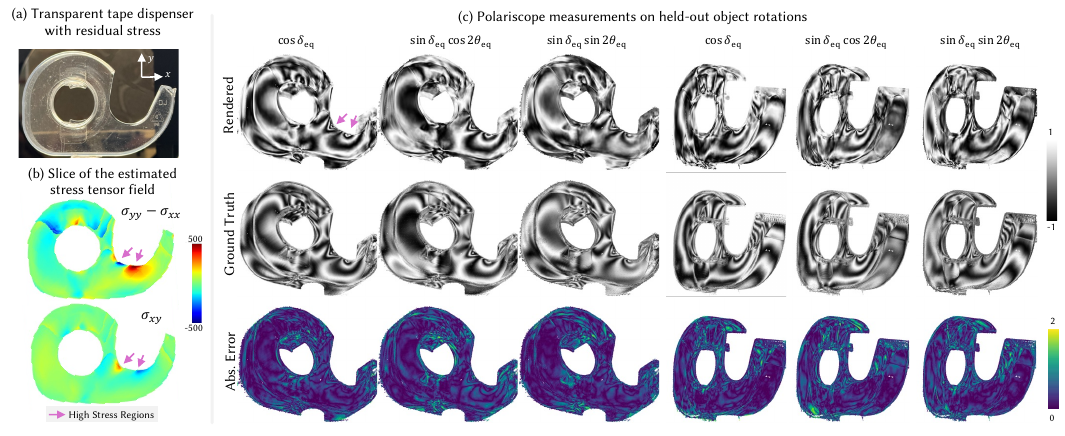}
    \caption{\textbf{NeST Application: Novel View Stress Visualization} NeST enables visualizing the 3D stress tensor field distribution within objects by rendering photoelastic fringes for novel views. From multi-axis polariscope measurements of a transparent tape dispenser (a), NeST estimates the underlying stress field (b) and can render fringes matching held-out test rotations (c), allowing interactive visualization of stress distributions.}
    \label{fig:app_visualization}
\end{figure*}

\paragraph{Novel View Stress Visualization}
NeST enables new ways to visualize the 3D stress tensor field distribution within common objects. Objects such as a clear plastic tape dispenser (Fig.~\ref{fig:app_visualization}(a)) exhibit fringes when placed in a polariscope due to the residual stress distribution within these objects. The density of fringes is related to the amount of residual fringes and these fringes are a way to visualize the underlying stress tensor distribution \cite{bussler2015photoelasticity}. From a set of multi-axis polariscope measurements, we can estimate the underlying neural stress tensor field (b). The estimated stress reveals regions with high stress concentration, depicted as arrows in (b). Using our forward model, we can then render the measurements for new rotations of the object not seen during training. In Fig.~\ref{fig:app_visualization}(c), we demonstrate that these rendered measurements qualitatively match the captures using a held-out test set not used in the training. Thus NeST can enable visualizing the stress distribution in objects in an interactive manner by rendering photoelastic fringes as the object is viewed from novel views or rotations. 

\section{Discussion and Conclusions}
\label{sec:discussion}

In this paper, we have introduced a differentiable non-linear forward model for
3D photoelasticity, and integrated it into a neural stress
tensor tomography approach we call NeST. A novel multi-axis polariscope setup is developed to capture the required measurements. We have demonstrated the efficacy
of the NeST on complex shapes in simulation, and on simpler
geometries in real experiments.

To bridge the complexity gap between simulation and experiment, the
primary challenge is to handle refraction and reflection at the object
surface. We see two fundamental approaches to tackle this issue, both
with their individual challenges:

\paragraph{Refractive index matching.} A hardware solution to surface
reflection and refraction would be to immerse the object in a liquid
of the same refractive index~\cite{trifonov2006tomographic}. The
primary challenge of this approach is that adding a liquid filled
container would significantly increase the complexity of the
multi-axis polariscope. It should also be noted that due to the
birefringence of the object, the refractive index matching would be
approximate. Still, this approach could likely reduce reflections and
refraction to a negligible amount.

\paragraph{Explicitly modeling boundary surfaces as Fresnel reflectors
and refractors.} An alternative approach would be to 3D scan the
object before stress analysis, and then explicitly compute ray
intersections with these boundary surfaces, applying Fresnel
laws. It might even be possible to attempt a joint estimation of the
shape and the stress field. This approach would not require any
hardware changes, however it significantly complicates the forward
model, since the ray paths in this model split into reflected and
refracted parts at every interface. Total internal reflectance would
further complicate the forward model.

Both of these approaches require significant new research
and expansions of the proposed approach, and are thus left for future
work. Nonetheless, we believe that our model and approach present a
significant advance in the 3D analysis of complex stress fields using
photoelasticity.

\bibliographystyle{ACM-Reference-Format}
\bibliography{references}
\appendix
\section{Additional derivations for the 3D photoelasticity image formation model}
\subsection{Derivation of principal stress} 
\label{app:principal_stress}
We are given the $2\times2$ projection $\mathbf{S}'$ of a Cartesian stress tensor $\mathbf{S}$ along the plane spanned by the orthonormal basis vectors $\mathbf{u}, \mathbf{v}$.
\begin{equation}
   \mathbf{S}' = \begin{bmatrix} \mathbf{u}^T \mathbf{S}\mathbf{u} & \mathbf{u}^T\mathbf{S}\mathbf{v} \\ \mathbf{u}^T\mathbf{S}\mathbf{v} & \mathbf{v}^T\mathbf{S}\mathbf{v} \end{bmatrix} 
   \triangleq \begin{bmatrix}
      \sigma_{\mathbf{u}\mathbf{v}} & \sigma_{\mathbf{u}\mathbf{v}} \\
      \sigma_{\mathbf{uv}} & \sigma_{\mathbf{vv}}
   \end{bmatrix} \; . 
   \label{eq:app_projected_stress}
\end{equation}
We aim to find basis vectors $\mathbf{w_1}, \mathbf{w_2}$ in the same space spanned by $\mathbf{u}, \mathbf{v}$ such that the projected stress tensor corresponding to this new basis $\mathbf{S}''$ is diagonal (i.e. only normal stress and no tangential stress):
\begin{equation}
 \begin{bmatrix} \mathbf{w}_1^T \mathbf{S}\mathbf{w}_1 & \mathbf{w}_1^T\mathbf{S}\mathbf{w}_2 \\ \mathbf{w}_1^T\mathbf{S}\mathbf{w}_2 & \mathbf{w}_2^T\mathbf{S}\mathbf{w}_2 \end{bmatrix} 
   \triangleq \mathbf{S}'' \triangleq \begin{bmatrix}
      \sigma_1 & 0 \\
      0 & \sigma_2
   \end{bmatrix} \;. 
   \label{eq:app_principal_stress}
\end{equation}
We denote the two sets of basis vectors with matrices as, \begin{align}
\mathbf{U} &= \begin{bmatrix}
  | & | \\
  \mathbf{u} & \mathbf{v} \\
  | & |
\end{bmatrix} &
\mathbf{P} &= \begin{bmatrix}
  | & | \\
  \mathbf{w}_1 & \mathbf{w}_2 \\
  | & |
\end{bmatrix} \; .
\label{eq:P_v1_v2}
\end{align}
We can then rewrite Eq~\ref{eq:app_projected_stress}  and  \ref{eq:app_principal_stress} as : \begin{align}
   \mathbf{S}' &= \mathbf{U}^T
                \mathbf{S}
                \mathbf{U}
                \label{eq:app_projected_stress_2}\\
   \mathbf{S}'' &= 
                \mathbf{P}^T
                \mathbf{S}
                \mathbf{P}
                \label{eq:app_principal_stress_2}\;.
\end{align}
As $\mathbf{w_1}$, $\mathbf{w_2}$ lie in the same space spanned by $\mathbf{u}$,$\mathbf{v}$ we can relate $\mathbf{w}_1, \mathbf{w}_2$  to $\mathbf{u},\mathbf{v}$ by an orthogonal transformation matrix $\mathbf{T}$ as :
\begin{align}
\mathbf{P}
    =  
    \mathbf{U}\mathbf{T}
    \label{eq:PUT} \;.
\end{align}
Substituting this equation to Eq.~\ref{eq:app_principal_stress_2},
\begin{equation}
    \mathbf{S}'' =  \mathbf{T}^T\mathbf{S}'\mathbf{T} = \begin{bmatrix}
      \sigma_1 & 0 \\
      0 & \sigma_2
   \end{bmatrix} \; .
   \label{eq:TST}
\end{equation}
To diagonalize $\mathbf{S}'$, $\mathbf{T}$ should correspond to the eigenvectors of $\mathbf{S}'$. 

As $\mathbf{T}$ is orthogonal, we can parametrize it with $\theta$ as :
\begin{align}
    \mathbf{T} = \begin{bmatrix}
        \cos \theta & \sin \theta \\
        \sin \theta & \cos \theta
    \end{bmatrix} \; .
\end{align}
Substituting above equation into Eq~\ref{eq:TST}  and comparing the elements of the left and right hand side matrices, we get the following set of equations:
\begin{align}
    \sigma_\mathbf{uu} \cos^2 \theta - 2\sigma_\mathbf{uv} \cos \theta \sin \theta + \sigma_\mathbf{vv} \sin^2 \theta 
   &= \sigma_1 
    \label{eq: sigma_1}\\
    (\sigma_\mathbf{uu} - \sigma_\mathbf{vv})\sin \theta \cos \theta + \sigma_\mathbf{uv} (\cos^2 \theta - \sin^2 \theta)
    &= 0\label{eq:zero_relation}\\
    \sigma_\mathbf{uu} \sin^2 \theta + 2\sigma_\mathbf{uv} \cos \theta \sin \theta + \sigma_\mathbf{vv} \cos^2 \theta &= \sigma_2 \label{eq: sigma_2}\\
\end{align}
From Eq \ref{eq:zero_relation} and using trigonometric relations,
\begin{align}
    \tan 2 \theta &= \frac{2 \sigma_{\mathbf{uv}}}{\sigma_{\mathbf{vv}}- \sigma_{\mathbf{uu}}}\\
    \theta &= \frac{1}{2}\tan^{-1}\left(\frac{2\sigma_{\mathbf{uv}}}{\sigma_{\mathbf{vv}}-\sigma_{\mathbf{uu}}}\right) \;. \label{eq:app_theta} 
\end{align}
From Eq~\ref{eq:PUT} and Eq~\ref{eq:P_v1_v2}, 
\begin{align}
    \mathbf{w}_1 &= \mathbf{u} \cos \theta + \mathbf{v} \sin \theta &
    \mathbf{w}_2 &= \mathbf{u} \sin \theta + \mathbf{v} \cos \theta \;.
\end{align}
Thus, $\theta$ is the angle made by $\mathbf{w}_1$ and matches the definition in the main paper. 
Subtracting Eq.~\ref{eq: sigma_1} from Eq~\ref{eq: sigma_1}  and using trigonometric relations, we have:
\begin{align}
   (\sigma_\mathbf{vv}-\sigma_\mathbf{uu})\cos 2\theta + 2\sigma_\mathbf{uv} \sin 2\theta = \sigma_2 - \sigma_1
\end{align}   
Susbtituting value of theta from Eq.~\ref{eq:app_theta} and simplifying we get,
\begin{align}
   \sigma_2 - \sigma_1 = \sqrt{\left(\sigma_{\mathbf{vv}} - \sigma_{\mathbf{uu}} \right)^2 + {\sigma_{\mathbf{uv}}}^2} \; .
\end{align}
\subsection{Derivation of our 3D photoelasticity formulation from integrated photoelasticity equation}
In Sec~4, we used the principal stress directions and values to derive our approximate 3D photoelasticity forward model. Here we derive our formulation from the more general stress optic relation as demonstrated in prior works in integrated photoelasticity \cite{aben1979integrated,bussler2015photoelasticity} and show how our forward model is an approximation. We will later use this model for deriving the first-order/linear approximation model.  
This projected stress tensor $\mathbf{S}'(t)$ induces a weak birefringence along the ray at $\mathbf{p}(t)$ \cite{aben1979integrated}. This birefringence effect can be modeled as the change in the Jones vector at that point through the stress-optic relation
\begin{equation}
    \frac{d\mathbf{E}(t)}{dt} = \mathbf{G}(t)\mathbf{E}(t) \;,
    \label{eq:diff_stress_optic}
\end{equation}
where $\mathbf{G}(t)$ depends on the projected stress as
\begin{equation}
\mathbf{G}(t) = -\frac{i 2\pi C}{\lambda} \begin{bmatrix} \frac{1}{2} \left(s'_{11}(t) - s'_{22}(t)\right) & s'_{12}(t) \\ s'_{12}(t) & -\frac{1}{2} \left(s'_{11}(t) - s'_{22}(t)\right) \end{bmatrix} .
\end{equation}
Here $C$ is the stress-optic coefficient that depends on the material properties of the object and $\lambda$ is the wavelength of light. 

By defining
$\sigma(t) = \frac{1}{2}\left(s'_{11}(t) - s'_{22}(t)\right)$ and $\tau(t) = s'_{12}(t)$ we can denote $\mathbf{G}(t)$ as 
\begin{align}
\mathbf{G}(t) = 
-i \frac{2\pi C}{\lambda}
\begin{bmatrix}
\sigma(t) & \tau(t) \\
\tau(t) & -\sigma(t) \;.
\end{bmatrix} \label{eq:G_sig_tau}
\end{align}
and the stress optic relation from Eq.~\ref{eq:diff_stress_optic} can be written as
\begin{equation}
\frac{dE(t)}{dt} = -i \frac{2\pi C}{\lambda}
\begin{bmatrix}
\sigma(t) & \tau(t) \\
\tau(t) & -\sigma(t) \;.
\end{bmatrix}
E(t) \label{eq:stress_optic_sig_tau}
\end{equation}



Integrated photoelasticity \cite{aben1979integrated} aggregates the change in Jones vector at each point along the ray from Eq.~\ref{eq:stress_optic_sig_tau} aggregated to obtain the net change of Jones vector along the ray as
\begin{equation}
    \int_{t_n}^{t_f}{\frac{d\mathbf{E}(t)}{dt}} = \int_{t_n}^{t_f}{\mathbf{G}(t)\mathbf{E}(t)} \;,
    \label{eq:integrated_stress_optic}
\end{equation}
Bu{\ss}ler et al. \cite{bussler2015photoelasticity} compute this integral as Euler integration using the 4th order Runge Kutta (RK-4) algorithm. However, this approach is not well suited for solving the inverse optimization. Here we derive the Monte-Carlo integration formulation for integrated photoelasticity and show how Poincar\'{e}'s equivalence theorem \cite{hammer2004characteristic} simplifies the integral computation.
\paragraph{Monte-Carlo Integration}
Consider samples \( t_i \) along the ray \( j = 1, 2, \ldots, N \). Assuming \( G \) is constant between the samples \( t_i \) and \( t_{i+1} \), integrated photoelasticity equation Eq. ~\ref{eq:integrated_stress_optic} between $t_i$ and $t_{i+1}$ can be approximated as

\begin{equation}
\int_{t_i}^{t_{i+1}} d\mathbf{E(t)} = \mathbf{G}(t_i) \int_{t_i}^{t_{i+1}} \mathbf{E(t)}dt
    \label{eq:integrated_stress_optic}
\end{equation}


\begin{equation}
\mathbf{E}(t_{i+1}) = \text{Exp}\left(\mathbf{G}(t_i) \Delta t_i \right) \mathbf{E}(t_i)
\end{equation}
where Exp is the complex matrix exponential function and  \( \Delta t_i = t_{i+1} - t_i \). We denote the transformation from $\mathbf{E}(t_i)$ to $\mathbf{E}(t_{i+1})$ as the Jones matrix $\mathbf{J}(t_i)$:
\begin{align}
    \mathbf{J}(t_i) = \text{Exp}\left(\mathbf{G}(t_i) \Delta t_i \right) 
\end{align}
We can then express this Jones matrix as a function of projected stress at $t_i$ by substituting $\mathbf{G}(t)$ from Eq.~\ref{eq:G_sig_tau}:
\begin{align}
    \mathbf{J}(t_i) = \text{Exp}\left(-i \frac{2\pi C}{\lambda}
\begin{bmatrix}
\sigma(t) & \tau(t) \\
\tau(t) & -\sigma(t) 
\end{bmatrix} \Delta t_i \right)  
\label{eq:J_G}\;.
\end{align}
In App.~\ref{app:exp_j_A}, we show that Eq.~\ref{eq:J_G} can be simplified as,
\begin{equation}
\mathbf{J}(t_i) = \cos \left( -\frac{2\pi C}{\lambda} \sqrt{\sigma_{i}^2 + \tau_{i}^2} \right) \mathbf{I} 
+ I \frac{\sin \left( -\frac{2\pi C}{\lambda} \sqrt{\sigma_{i}^2 + \tau_{i}^2} \right)}{\sqrt{\sigma_{i}^2 + \tau_{i}^2}} G_{i}
\end{equation}
where we denote $\sigma(t_i)$ and $\tau(t_i)$ as $\sigma_i$ and $t_i$ respectively.
We define the parameters $\delta_i$ and $\theta_i$ that are termed in the photoelastic literature as isochromatic and isoclinic parameters \cite{ramesh2021developments}.
\begin{align}
\delta_i = -\frac{4\pi C}{\lambda} \sqrt{\sigma_i^2 + \tau_i^2} \quad \quad 
\theta_i = \frac{1}{2} \tan^{-1} \left( \frac{\sigma_{i}}{\tau_{i}} \right)
\label{eq:delta_theta}
\end{align}
We can express the Jones matrix in terms of these parameters as,
\begin{equation}
\mathbf{J}\left(t_i\right) = \cos \delta_i/2
\begin{bmatrix}
1 & 0 \\
0 & 1
\end{bmatrix}
+ i \sin \delta_i/2
\begin{bmatrix}
\cos 2\theta_i & \sin 2\theta_i \\
\sin 2\theta_i & -\cos 2\theta_i
\end{bmatrix}
\label{eq:J_retarder} \; ,
\end{equation} 
This Jones matrix can also be understood as that of a retarder element with retardation $\delta$ oriented with slow axis making an angle $\theta$ with the `horizontal` basis vector $\textbf{u}$ \cite{collett2005field}.

\subsubsection{Derivation of \( \exp(jA) \) for a Symmetric, Trace-Free 2x2 Matrix}
\label{app:exp_j_A}
Consider a symmetric, trace-free 2x2 matrix \( A \) defined as
\[
A = \begin{pmatrix}
\sigma & \tau \\
\tau & -\sigma
\end{pmatrix}.
\]

We aim to find the matrix exponential \( \exp(jA) \) using the Taylor series expansion for the matrix exponential, given by
\[
\exp(X) = I + X + \frac{X^2}{2!} + \frac{X^3}{3!} + \cdots,
\]
where \( I \) is the identity matrix.


First, we find \( A^2 \) as follows:
\[
A^2 = \begin{pmatrix}
\sigma & \tau \\
\tau & -\sigma
\end{pmatrix}
\begin{pmatrix}
\sigma & \tau \\
\tau & -\sigma
\end{pmatrix}
= \begin{pmatrix}
\sigma^2 + \tau^2 & 0 \\
0 & \sigma^2 + \tau^2
\end{pmatrix}.
\]


The Taylor series for \( \exp(jA) \) can be separated into real and imaginary parts as follows:

Real Part:
\[
I - \frac{A^2}{2!} + \frac{A^4}{4!} - \cdots
\]

Imaginary Part:
\[
jA - j\frac{A^3}{3!} + j\frac{A^5}{5!} - \cdots
\]


Notice that higher powers of \( A \) cycle through combinations of \( I \), \( A \), and \( A^2 \). For example, \( A^3 = A(A^2) = A(\sigma^2 + \tau^2) \) and so on. This pattern allows us to simplify the series into sums involving \( I \), \( A \), and \( A^2 \).

Applying Euler's formula \( \exp(jx) = \cos(x) + j\sin(x) \), the real and imaginary parts can be simplified as:

Real Part:
\[
\cos(\sqrt{\sigma^2 + \tau^2})I
\]

Imaginary Part:
\[
j \frac{\sin(\sqrt{\sigma^2 + \tau^2})}{\sqrt{\sigma^2 + \tau^2}} A
\]


Combining the real and imaginary parts, the final expression for \( \exp(jA) \) is:
\[
\exp(jA) = \cos(\sqrt{\sigma^2 + \tau^2})I + j \frac{\sin(\sqrt{\sigma^2 + \tau^2})}{\sqrt{\sigma^2 + \tau^2}} A.
\]
\subsection{First order Approximation of Integrated Photoelasticity}
\label{app:first_order_approx}
In Sec.~4, we derive that the effective Jones matrix along the ray can be approximated as the product of Jones matrices. Here, we show how the first-order approximation of this formulation leads to the linear tensor tomography formulation \cite{sharafutdinov2012integral} utilized by prior works \cite{szotten2011limited,lionheart2009reconstruction}.  

The effective Jones matrix is the matrix product of complex matrix exponentials, expressed as,
\begin{equation}
J_{\text{net}} = \prod_{i=1}^{N} \text{Exp} \left[ -i \frac{2\pi C}{\lambda} G(t_i) \Delta t_i \right] E(t_0)
\end{equation}
Using Baker-Campbell-Hausdorff (BCH) formula \cite{hall2000elementary}, we can express the logarithm of the product of two exponentials of operators (or matrices) as a single exponential. For non-commuting matrices \(A\) and \(B\), it's given by:
\begin{align}
\log(\text{Exp}(A) \text{Exp}(B)) &= A + B + \frac{1}{2}[A, B] + \frac{1}{12}([A, [A, B]] + [B, [B, A]]) \notag \\
& - \frac{1}{24}[B, [A, [A, B]]] + \cdots 
\end{align}
where \([A, B] = AB - BA\) is the commutator of \(A\) and \(B\).

Thus under first-order approximation, we can approximate the product of complex exponentials in Eq~\ as the exponential of the sum of matrices
\begin{equation}
J_{\text{net}} = \text{Exp} \left[ -i \frac{2\pi C}{\lambda} \sum_{i=1}^{N} G(t_i) \Delta t_i \right]
\end{equation}

We denote the combined matrix \( G_{\text{net}} \) as:
\begin{equation}
G_{\text{net}} = \sum_{i=1}^{N} G(t_i) \Delta t_i = 
\begin{bmatrix}
\sigma_{\text{net}} & \tau_{\text{net}} \\
\tau_{\text{net}} & -\sigma_{\text{net}}
\end{bmatrix}
\end{equation}

Where
\begin{equation}
\sigma_{\text{net}} = \sum_{i=1}^{N} \sigma_i \Delta t_i
\end{equation}
\begin{equation}
\tau_{\text{net}} = \sum_{i=1}^{N} \tau_i \Delta t_i
\end{equation}

\begin{equation}
J_{\text{net}} = \text{Exp}\left[ -i \frac{2\pi C}{\lambda} G_{\text{net}} \right]
\end{equation}

Using derivation in App.~\ref{app:exp_j_A}, this can be simplified as
\begin{equation}
J_{\text{net}} = \cos \left( -\frac{2\pi C}{\lambda} \sqrt{\sigma_{\text{net}}^2 + \tau_{\text{net}}^2} \right) \mathbf{I} 
+ j \frac{\sin \left( -\frac{2\pi C}{\lambda} \sqrt{\sigma_{\text{net}}^2 + \tau_{\text{net}}^2} \right)}{\sqrt{\sigma_{\text{net}}^2 + \tau_{\text{net}}^2}} G_{\text{net}}
\end{equation}

We define the parameters $\delta$ and $\theta$ that are termed in the photoelastic literature as isochromatic and isoclinic parameters \cite{ramesh2021developments}.
\begin{equation}
\delta = -\frac{4\pi C}{\lambda} \sqrt{\sigma_{\text{net}}^2 + \tau_{\text{net}}^2}
\label{eq:delta}
\end{equation}
\begin{equation}
2\theta = \tan^{-1} \left( \frac{\sigma_{\text{net}}}{\tau_{\text{net}}} \right)
\label{eq:theta}
\end{equation}

\begin{equation}
J_{\text{net}} = \cos \delta \mathbf{I} + j \sin \delta
\begin{bmatrix}
\cos \theta & \sin \theta \\
\sin \theta & -\cos \theta
\end{bmatrix}
\label{eq:J_net}
\end{equation}

\subsection{Expressions for all the captured polariscope measurements}
In Table ~\ref{table:all_intensity_expressions}, we provide expressions for all the $16$ measurements we capture with our acquisition setup for each object rotation. As described in Sec.~7.1, we capture raw measurements with the polarization camera by varying QWP1 ($\beta_1$) and QWP2($\beta_2$) rotations with LP1 ($\alpha_1$) set to 90 degrees. We derive the expressions for the captured intensity $I_\text{cap}$ as a function of both the unique elements in the Jones matrix $\left(a_\text{eq},b_\text{eq},c_\text{eq},d_\text{eq}\right)$ and the characteristic parameters $\left(\delta_\text{eq},\theta_\text{eq},\gamma_\text{eq}\right)$. We drop the subscript,eq, in the table for simplicity. $\gamma' = 2\gamma - \theta$. From Tab.~\ref{table:all_intensity_expressions}, it is evident that all the measurements are a function of 6 unique expressions $\tilde{I}_1$ to $\tilde{I}_6$ as described in Sec.~7.1.

\begin{table}[h]
\small
\centering
\begin{tabular}{@{}l@{\hskip 4pt}c@{\hskip 2pt}c@{\hskip 2pt}c@{\hskip 4pt}lr@{}} 
\toprule
$I_i$ & \textbf{$\rho$ } & \textbf{$\eta$ } & \textbf{$\beta$ } & $2I_\text{cap}(\delta,\theta,\gamma)$ & $I_\text{cap}(a,b,c,d)$ \\
\midrule
$I_0$    & $45^\circ$ & $0^\circ$     & $0^\circ$       & $1 -  \sin \delta \sin 2 \theta $                & $ 1 - 2ac + 2bd $\\
$I_1$    & $45^\circ$ & $0^\circ$     & $45^\circ$   & $1 -  \cos \delta$                                         & $ 2b^2 + 2c^2 $\\
$I_2$    & $45^\circ$ & $0^\circ$     & $90^\circ$   & $1 +  \sin \delta \sin 2 \theta $                & $ 1 + 2ac - 2bd $\\
$I_3$    & $45^\circ$ & $0^\circ$     & $135^\circ$  & $1 +  \cos \delta $                                        & $ 2a^2 + 2d^2 $\\
$I_4$    & $45^\circ$ & $45^\circ$    & $0$       & $1 +  \cos \delta $                                        & $ 2a^2 + 2d^2 $\\
$I_5$    & $45^\circ$ & $45^\circ$    & $45^\circ$   & $1 +  \sin \delta \cos 2 \theta$                & $ 1 + 2ab + 2cd $\\
$I_6$    & $45^\circ$ & $45^\circ$    & $90^\circ$   & $1 -  \cos \delta$                                         & $ 2b^2 + 2c^2 $\\
$I_7$    & $45^\circ$ & $45^\circ$    & $135^\circ$  & $1 -  \sin \delta \cos 2 \theta $                & $ 1 - 2ab -2cd $\\
$I_8$    & $90^\circ$ & $0^\circ$     & $0^\circ$       & $1 -  \cos 2 \gamma' \cos 2 \theta-  \sin 2 \gamma' \sin 2 \theta \cos \delta $                        & $ 2c^2 + 2d^2  $\\
$I_9$    & $90^\circ$ & $0^\circ$     & $45^\circ$  & $1 +  \sin \delta \sin 2 \gamma'$                & $ 1 + 2ac + 2bd    $\\
$I_{10}$ & $90^\circ$ & $0^\circ$     & $90^\circ$ & $1 +  \cos 2 \gamma' \cos 2 \theta+  \sin 2 \gamma' \sin 2 \theta \cos \delta$         & $ 2a^2 + 2b^2  $\\
$I_{11}$ & $90^\circ$ & $0^\circ$     & $135^\circ$& $1 -  \sin \delta \sin 2 \gamma'$                & $ 1 - 2ac - 2bd  $\\
$I_{12}$ & $90^\circ$ & $45^\circ$    & $0^\circ$     & $1 -  \sin \delta \sin 2 \gamma'$                & $ 1 - 2ac - 2bd  $\\
$I_{13}$ & $90^\circ$ & $45^\circ$    & $45^\circ$  & $1 -  \cos 2 \gamma' \sin 2 \theta + \cos 2 \theta \sin 2 \gamma' \cos \delta $         & $ 1 + 2ad - 2bc $\\
$I_{14}$ & $90^\circ$ & $45^\circ$    & $90^\circ$ & $1 +  \sin \delta \sin 2 \gamma'$                & $ 1 + 2ac + 2bd $\\
$I_{15}$ & $90^\circ$ & $45^\circ$    & $135^\circ$& $1 +  \cos 2 \gamma' \sin 2 \theta-  \cos 2\theta \sin 2\gamma' \cos \delta$         & $ 1 - 2ad + 2bc $\\
\bottomrule
\end{tabular}
\caption{Expressions for captured intensity measurements as a function of both unique elements of the equivalent Jones matrix and the characteristic parameters of the equivalent Jones matrix. The subscript, eq, is dropped for simplicity. }
\label{table:all_intensity_expressions}
\end{table}

\section{Capture setup details}
\subsection{Calibration of polariscope angles}

Our polariscope setup involves mechanical rotations of the illumination linear polarizer LP1 and quarter-wave plate QWP1 and motorized rotation of the camera quarter-wave plate QWP2. The camera linear polarizer LP2 has fixed rotations at four angles: 0, 45, 90 and 135 degrees because LP2 corresponds to a polarizer grid within the linear polarizer camera that is placed directly on top of the sensor. In our setup, we need to ensure that the rotations for all the polariscope elements are defined based on the same reference coordinate system. We consider the coordinate system of LP1 as the fixed coordinate system and sequentially calibrate the rotations for LP2, QWP1, and QWP2 based on this coordinate system. By considering no specimen and only free space between the illumination and the camera, we utilize analytical expressions of the obtained intensity as a function of the polariscope element angles to perform this sequential calibration.

\subsection{Calibration of multi-axis rotation}

In order to calculate the pose of the camera relative to the object, a calibration process is required. Both the lens intrinsics and extrinsics are unknown, so we must calibrate both. We rely on COLMAP \cite{schoenberger2016colmap} to calculate these. The procedure is performed in three steps:

\begin{enumerate}
    \item Complete a capture of a high-contrast object with the full azimuth and elevation ranges, respectively.
    \item Run COLMAP \cite{schoenberger2016colmap} on the acquired images.
    \item Fit a regularly-spaced grid to the pose estimates.
\end{enumerate}

The regularly-spaced grid assumes the relative pose between the camera and object remains fixed, and so the pose offset from this grid is the calibrated pose.

\subsection{Component List}
Table \ref{tab:components_list} contains a list of components used to build our multi-axis polariscope experimental setup (Sec.~7.1 ).

\begin{table}[h]
\centering
\begin{tabular}{|c|c|c|}
\hline
\textbf{Component} & \textbf{Vendor} & \textbf{Part number} \\
\hline
LED Panel& NEEWER& PT-1765\\
QWP1& Izgut& Raw3D\\
Rotation Stage (x2)& Thorlabs& HDR50\\
Bandpass Filter& Thorlabs& FLH532-10\\
QWP2& Newport& 10RP44-1\\
QWP2 Rotation Mount& Thorlabs& ELL14\\
DSLR Lens& Nikon& NIKKOR 50mm f1.4\\
Polarization Camera& LUCID& PHX050S-OC\\
\hline
\end{tabular}
\caption{Various components used to build the prototype.}
\label{tab:components_list}
\end{table}


\subsection{Object Masking}
We mask out the background from the captured measurements for computing rendering loss and for visualization purposes. The background pixels corresponds to either the light rays that pass from illumination to the detection without encountering the object or the light rays that get occluded due to the object's mount. Occluded pixels have intensity close to zero and are thresholded out. We observe that the light rays that do not pass through the object have degree of linear polarization (DOLP) close to 1. We compute the degree of polarization (DOLP) from all the raw measurements captured at the polarization camera per object rotation and use the minimum DoLP to mask out the background pixels passing from free space.

\begin{figure}[h]
    \centering
    \includegraphics[width=1\linewidth]{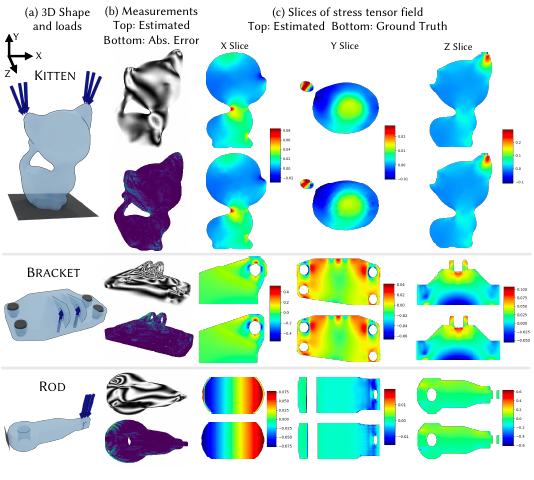}
    \caption{\textbf{Evaluation on additional simulated datasets.} We evaluate the accuracy of NeST on estimating stress tensor fields and on rendering polariscope measurement $cos \delta_\text{eq}$ on additional datasets created using the 3D-TSV stress field dataset \cite{wang_mechanical_2017}. }
    \label{fig:supp_3dtsv}
\end{figure}

\begin{figure}[h]
    \centering
    \includegraphics[width=1\linewidth]{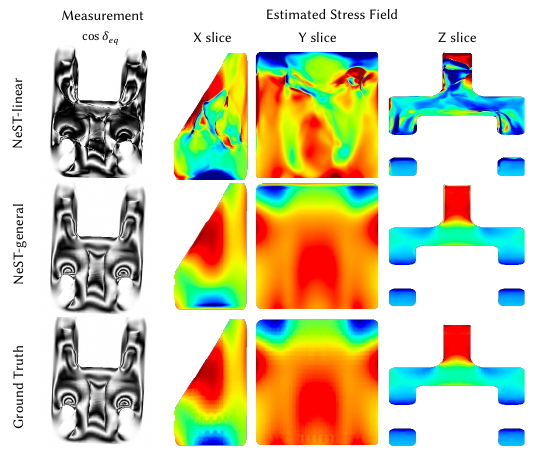}
    \caption{\textbf{\texttt{NeST-linear} vs \texttt{NeST-general} on simulated dataset.} By utilizing the more general non-linear forward model, \texttt{NeST-general} fits better to the complex photoelastic fringes for the \textsc{Bearing} dataset and the estimated stress field is closer to the ground truth compared to using the more approximate linear forward model. }
    \label{fig:supp_gen_vs_lin}
\end{figure}


\begin{figure}[h]
    \centering
    \includegraphics[width=1\linewidth]{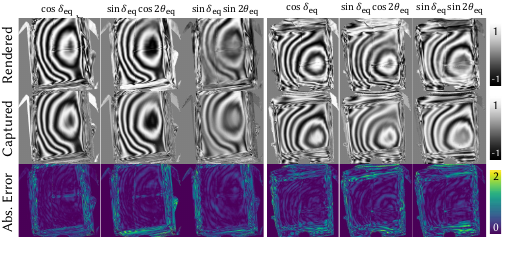}
    \caption{\textbf{Evaluation of polariscope measurements rendered by NeST for additional real object.} We estimate the underlying residual stress field in a plastic box using NeST and show that the polariscope measurements rendered on held out rotations match the ground truth.}
    \label{fig:supp_app_viz}
\end{figure}
\end{document}